\pdfoutput=1
\documentclass[logo]{sofa_lab_2026}

\usepackage[numbers,sort&compress]{natbib}
\usepackage{lineno}
\usepackage{xurl}
\usepackage{mathtools}
\usepackage{nicefrac}

\usepackage{subcaption}
\usepackage{array}
\usepackage{multirow}
\usepackage[normalem]{ulem}
\usepackage{algorithm}
\usepackage{algorithmic}

\usepackage[capitalize,noabbrev]{cleveref}

\newcommand{\cmark}{\ding{51}}
\newcommand{\xmark}{\ding{55}}

\newcommand{\rom}{ROM}
\newcommand{\romcsc}{ROM$_{\text{CSC}}$}
\newcommand{\ci}[1]{{\tiny$\pm$#1}}

\newcommand{\rev}[1]{#1}
\newcommand{\revhead}[1]{#1}
\newcommand{\revnextfloat}{}
\newcommand{\revon}{}

\newcommand{\transcriptcaption}[1]{%
  \begin{center}%
  \captionsetup{hypcap=false}%
  \setlength{\abovecaptionskip}{0.2cm}%
  \captionof{table}{#1}%
  \end{center}%
  \vspace{-0.8em}%
}
\newenvironment{transcriptbox}[1]{%
  \begin{tcolorbox}[
    enhanced,
    breakable,
    colback=gray!10,
    colframe=black,
    arc=2mm,
    boxrule=0.6pt,
    left=4pt, right=4pt, top=5pt, bottom=5pt,
    before skip=5pt, after skip=8pt,
  ]%
  {\small\bfseries #1}\par\vspace{0.5em}%
}{%
  \end{tcolorbox}%
}

\title{ROM: Real-time Overthinking Mitigation via\\ Streaming Detection and Intervention}
\fancypagestyle{plain}{%
  \fancyhf{}%
  \fancyhead[C]{\footerfont ROM: Real-time Overthinking Mitigation}%
  \fancyfoot[R]{\footerfont\thepage}%
}
\hypersetup{%
  pdftitle={ROM: Real-time Overthinking Mitigation via Streaming Detection and Intervention},
  pdfauthor={Xinyan Wang, Xiaogeng Liu, Ming Pei, Chaowei Xiao}%
}

\author[1]{Xinyan Wang}
\author[2]{Xiaogeng Liu}
\author[1]{Ming Pei}
\author[2]{Chaowei Xiao}
\affil[1]{University of Wisconsin--Madison}
\affil[2]{Johns Hopkins University}

\renewcommand{\headerlogos}{%
  \headerlogo[38pt]{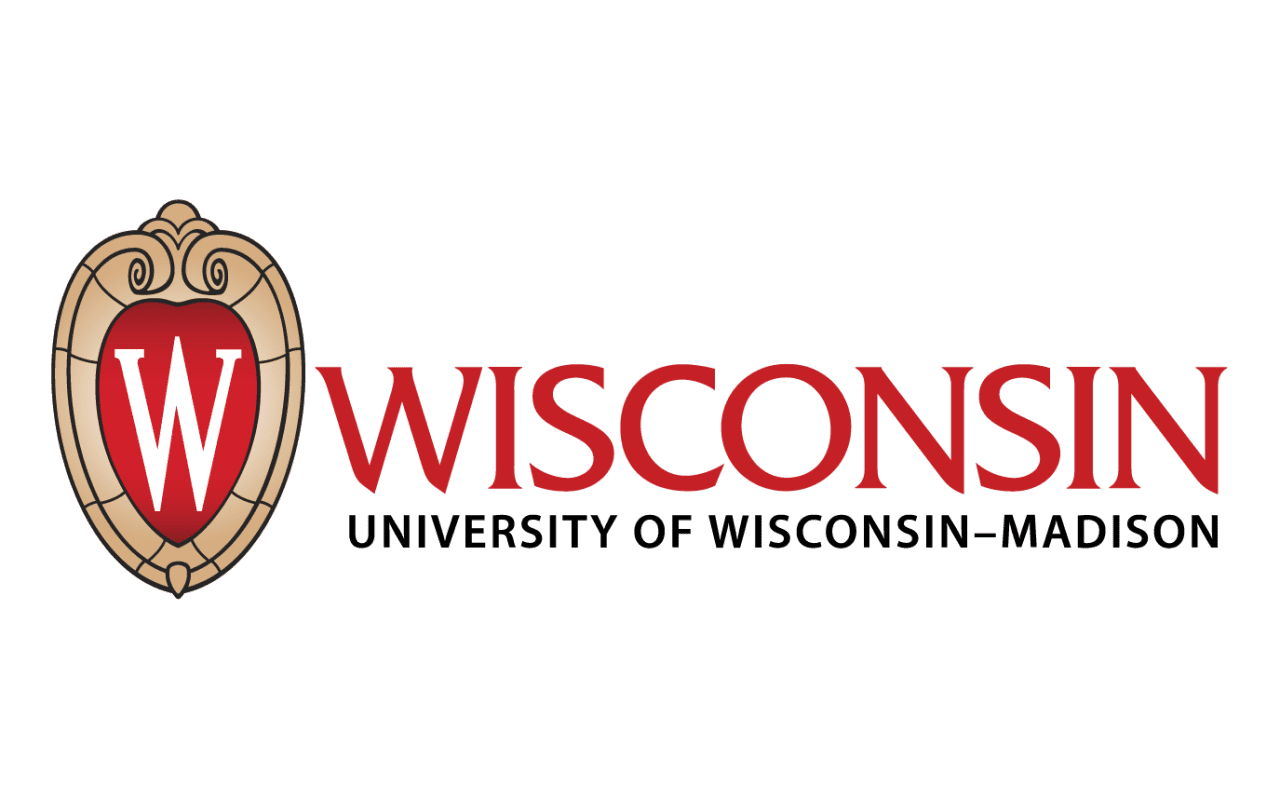}%
  \headerlogo[14pt]{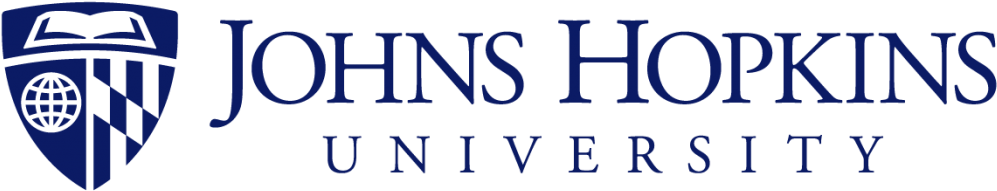}%
}

\newcommand{\projectbadge}[3]{%
  \href{#2}{%
    \raisebox{-0.2\height}{\includegraphics[height=1em]{#1}}\hspace{0.3em}%
    \texttt{#3}%
  }%
}
\newcommand{\projectbadges}{%
  \projectbadge{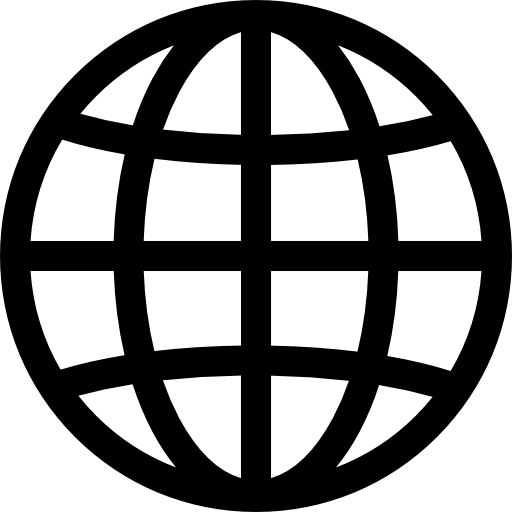}{https://xinyan-wang-stat.github.io/ROM-LRM/}{Website}%
  \hspace{0.6cm}%
  \projectbadge{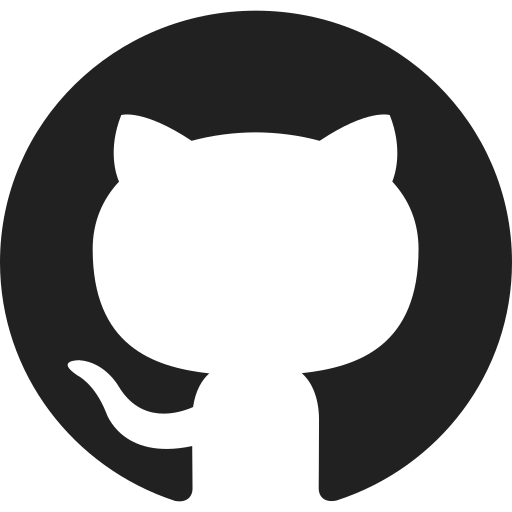}{https://github.com/SaFo-Lab/ROM}{Code}%
  \hspace{0.6cm}%
  \projectbadge{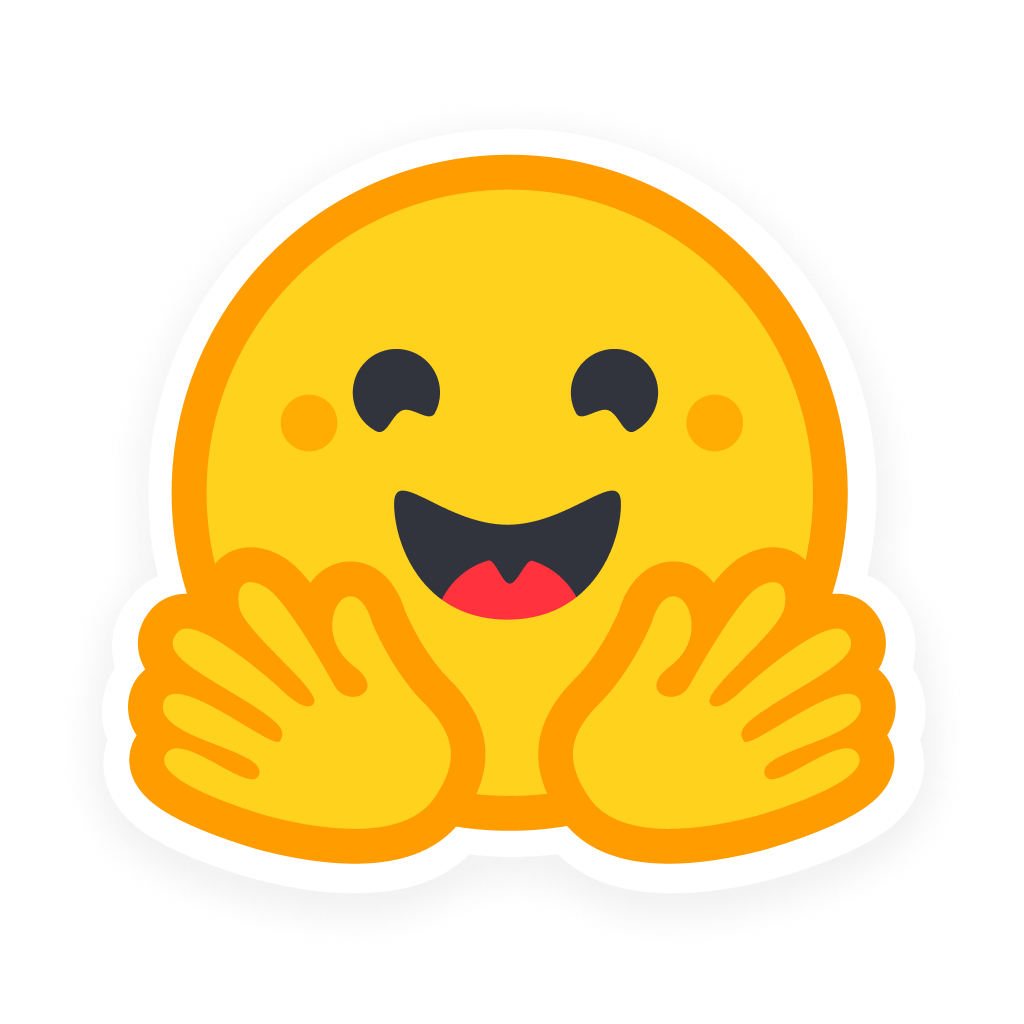}{https://huggingface.co/datasets/xinyan-wang/ROM}{Dataset}%
}

\begin{document}
\maketitle

\begin{abstract}
Large reasoning models (LRMs) often suffer from overthinking, continuing with redundant verification and repeated attempts even after reaching a correct solution. This wastes computation and can overturn an initially correct answer. We show that the transition from productive to redundant reasoning is decodable from late-layer hidden states. These representations separate the two phases around first-correct-solution boundaries beyond what position \rev{and lexical} controls explain, \rev{and distinguish correct from incorrect solution endings}. Building on this finding, we propose ROM (Real-time Overthinking Mitigation), which uses a lightweight streaming detector (\rev{0.07--0.25\%} of backbone parameters) to monitor a frozen LRM and intervene at well-formed reasoning boundaries without answer extraction or probe decoding. Counterfactual Self-Correction (CSC) augments detector training with wrong$\rightarrow$correct trajectories to preserve useful self-correction. \rev{Heads are trained separately for each backbone using shared MATH500 supervision and transfer zero-shot to four unseen benchmarks.} Evaluated against ten baselines across five backbones from three families and five benchmarks, ROM$_{\text{CSC}}$ reduces response length by 28--77\% (45\% on average), \rev{with significant reductions in all 25 settings and accuracy matching or exceeding vanilla decoding in 22}. \rev{Measurements on two backbones further show 20--70\% lower end-to-end latency with less than 0.5\% detector overhead per decoding step.}
\end{abstract}

\section{Introduction}
\label{sec:intro}

Large reasoning models (LRMs) such as OpenAI o1 and DeepSeek-R1~\citep{openai_o1,deepseek_r1} improve reasoning through long chain-of-thought (CoT) traces~\citep{wei2023chainofthoughtpromptingelicitsreasoning}. Yet they often suffer from \emph{overthinking}~\citep{chen2025think23overthinkingo1like, sui2025stopoverthinkingsurveyefficient}, continuing with redundant verification, repeated attempts, or unnecessary exploration after reaching a correct solution. This continuation increases computation and latency and can overturn an initially correct answer. The challenge is to identify when further reasoning becomes redundant while preserving the opportunity for useful self-correction.

Existing methods approximate reasoning saturation through manually chosen proxies, including backbone length objectives, intermediate-answer stability, confidence, entropy, and attention statistics~\citep{aggarwal2025l1controllinglongreasoning, fu2025efficientlyscalingllmreasoning, wang2025entropylangletextttthinkrangle}. Online detection must rely on the generated prefix alone, without access to reference answers or future reasoning. We therefore study overthinking as a transition in the reasoning process itself. We call the point at which a sufficient solution has been reached and further continuation becomes redundant the \emph{overthinking boundary}. Our central question is whether \textbf{this boundary leaves a detectable signature in hidden states \rev{beyond position or surface wording}, and whether that signature can guide online intervention}.

To investigate this question, we probe Qwen3-8B's late-layer hidden states around first-correct-solution boundaries on MATH500 (Sec.~\ref{sec:insight}). A linear probe distinguishes the adjacent productive and redundant phases with $85.9\%$ accuracy under response-level group cross-validation. The signal is localized at the boundary and persists under position controls. \rev{Lexical and matched-boundary controls further distinguish it from surface wording and generic solution-ending cues.} These findings motivate detecting the transition directly from hidden states.

Building on this finding, we introduce \textbf{ROM} (Real-time Overthinking Mitigation), to our knowledge the first method to directly supervise the productive-to-redundant transition. ROM attaches a lightweight streaming detector (\rev{0.07--0.25\%} of backbone parameters) to a frozen LRM. Offline answer verification identifies first-correct-solution boundaries to label the training traces. During decoding, the detector reads late-layer hidden states, integrates evidence over the prefix, and produces a token-level overthinking score without reference answers. When triggered, a boundary-aware policy backtraces to a sentence or solution boundary before eliciting a final answer. \rev{Each backbone's head is trained separately on the same labeled MATH500 traces.} ROM reuses the backbone's forward pass and requires no answer extraction or probe decoding at inference.

Early stopping must also distinguish redundant repetition from useful self-correction. Most labeled traces reach a correct answer on the first attempt, so a detector could learn attempt index as a shortcut and truncate later attempts that are still productive. We introduce \textbf{Counterfactual Self-Correction (CSC)}, which augments detector training with wrong$\rightarrow$correct trajectories and labels reasoning through the first correct solution as efficient. \rev{Our analyses show that CSC reduces reliance on attempt index and premature stopping on natural self-correction traces (Sec.~\ref{sec:ablations}).}

We evaluate the resulting approach against ten baselines across five backbones from three families and five benchmarks, \rev{with zero-shot transfer to the four benchmarks not used for training}. Our contributions are threefold.
\begin{itemize}[leftmargin=*,itemsep=1pt,topsep=2pt]
    \item \textbf{A hidden-state signal of overthinking.} We show that the productive-to-redundant transition is linearly decodable from late-layer representations, \rev{supported by lexical and matched-boundary controls in addition to position controls}.
    \item \textbf{Streaming detection and intervention.} We introduce ROM and CSC to control reasoning online with a frozen backbone while preserving useful self-correction.
    \item \textbf{Validation across models and tasks.} \romcsc{} reduces response length by 28--77\%, \rev{with significant reductions in all 25 settings and accuracy matching or exceeding vanilla in 22. End-to-end latency falls by 20--70\% on two backbones}.
\end{itemize}

\section{Related Work}
\label{sec:related}

\paragraph{Detecting and mitigating overthinking.}
Even simple problems can elicit long, repetitive continuations with little accuracy benefit~\citep{chen2025think23overthinkingo1like}. Model-based methods such as L1 and O1-Pruner~\citep{aggarwal2025l1controllinglongreasoning,luo2025o1prunerlengthharmonizingfinetuningo1like} optimize reasoning length through backbone training. Inference-time methods use a range of stopping signals. Answer-based approaches, including Dynasor/Certaindex~\citep{fu2025efficientlyscalingllmreasoning}, DEER~\citep{yang2025dynamicearlyexitreasoning}, PUMA-RD~\citep{min2026stop}, and PMA~\citep{yan2026pma}, rely on intermediate-answer extraction or probing (Sec.~\ref{sec:baseline_pareto}). Other approaches monitor entropy~\citep{wang2025entropylangletextttthinkrangle,guan2026rpdm}, termination-token attention or rank dynamics~\citep{li2026syncthinktrainingfreestrategyalign,wei2026evolutionthoughttrackingllm}, exit-neuron activations~\citep{liu2026neat}, or step-level embedding redundancy~\citep{sun2025refrain}. Learned hidden-state predictors assess candidate-answer correctness~\citep{zhang2025reasoningmodelsknowtheyre} or detect the earliest logical arrival of the model's eventual answer~\citep{nagle2026terminatorlearningoptimalexit}. A central challenge is distinguishing redundant continuation from useful self-correction: confidence may stabilize around an incorrect answer or remain unstable after a correct one (Sec.~\ref{sec:entropy_comparison}). ROM addresses this challenge through reference-verified first-correct-solution supervision and CSC augmentation, which explicitly labels reasoning through successful self-correction as productive. Its lightweight streaming detector and boundary-aware intervention reduce redundant reasoning without updating the backbone or requiring online answer extraction or additional probe decoding.

\paragraph{Latent-boundary control.}
ROM adapts streaming token-level monitoring, previously explored for safety and hallucination detection~\citep{sharma2025constitutionalclassifiersdefendinguniversal,li2025judgmentinterferenceearlystopping,xuan-etal-2025-shieldhead,krishna2025disentangledsafetyadaptersenable,obeso2025realtimedetectionhallucinatedentities,li2025kelpstreamingsafeguardlarge}, to overthinking mitigation. Offline boundary labels supervise a lightweight detector that integrates hidden-state evidence over the generated prefix into an online overthinking signal (Sec.~\ref{sec:labeling}). Detection and intervention operate at different granularities: the detector scores every token, while boundary-aware backtracing selects a well-formed reasoning boundary before eliciting the final answer. This allows ROM to monitor reasoning continuously while respecting its local structure when intervening. App.~Table~\ref{tab:three_properties} summarizes the distinctions from prior approaches.

\section{Method}
\label{sec:method}

\begin{figure*}[t]
    \centering
    \includegraphics[width=0.92\linewidth]{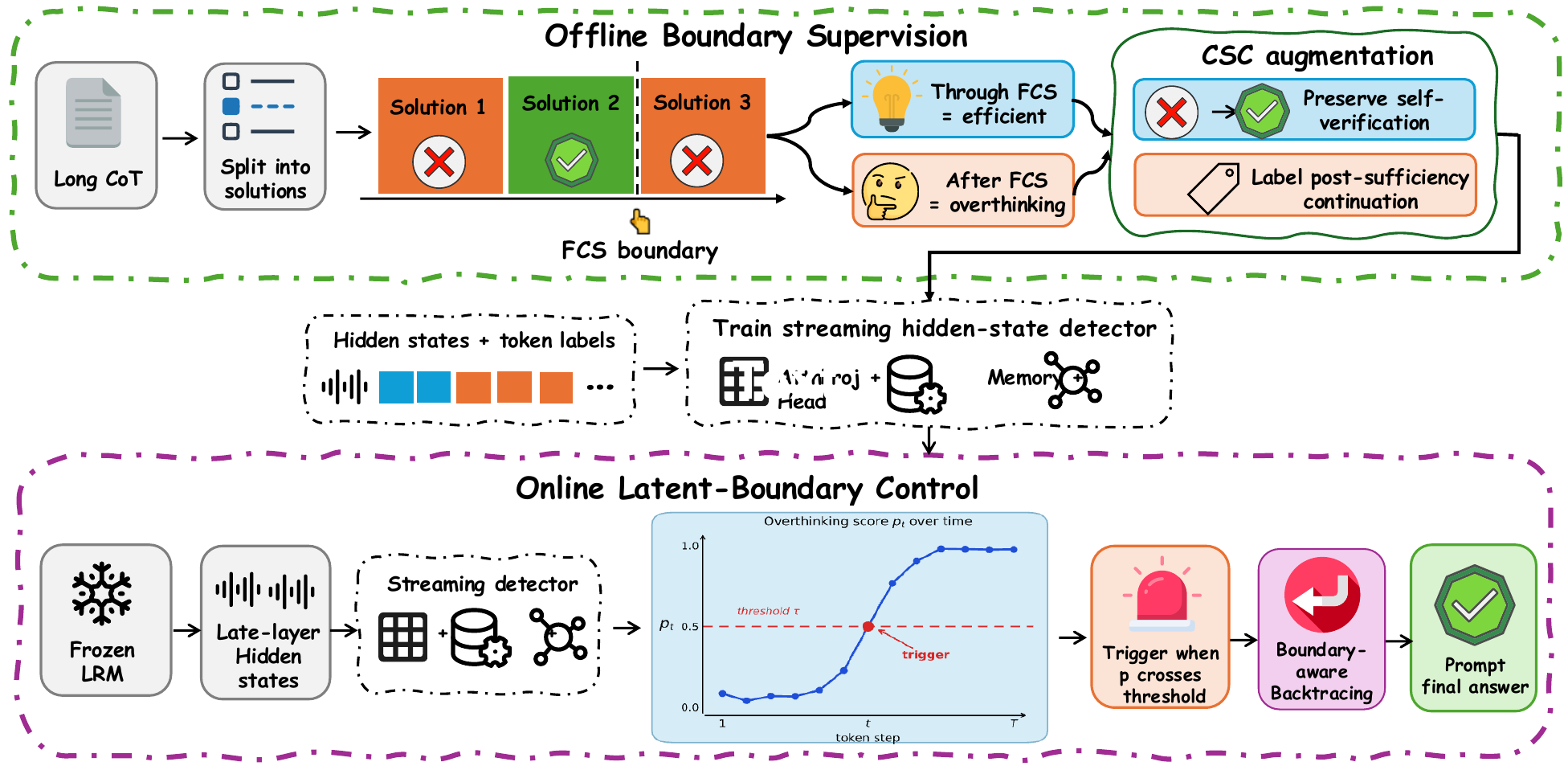}
    \caption{Overview of ROM. \textbf{Top}: attempt-level correctness identifies the first-correct-solution (FCS) boundary, yielding token-level labels. CSC adds wrong$\rightarrow$correct trajectories (Secs.~\ref{sec:labeling}--\ref{sec:csc}) to train the streaming detector (Sec.~\ref{sec:detection}). \textbf{Bottom}: during decoding, the detector triggers boundary-aware backtracing and final-answer forcing when its score $p_t$ crosses the threshold (Sec.~\ref{sec:intervention}).}
    \label{fig:framework}
\end{figure*}

We first show that the productive-to-redundant transition is decodable from hidden states (Sec.~\ref{sec:insight}), then use this signal for online control through a lightweight head on the frozen backbone (Fig.~\ref{fig:framework}).

\subsection{Motivation: A Latent Footprint of Overthinking}
\label{sec:insight}

\begin{figure}[t]
    \centering
    \begin{subfigure}[t]{0.38\linewidth}
        \centering
        \includegraphics[width=\linewidth]{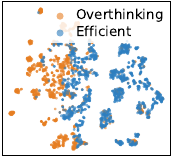}
        \caption{Boundary-token hidden states cluster by reasoning phase.}
        \label{fig:insight_tsne}
    \end{subfigure}
    \hfill
    \begin{subfigure}[t]{0.59\linewidth}
        \centering
        \includegraphics[width=\linewidth]{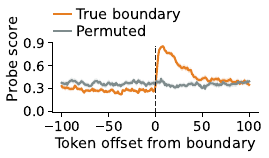}
        \caption{Probe scores rise at the true boundary but stay flat under permuted alignment.}
        \label{fig:aligned_probe_score}
    \end{subfigure}
    \caption{The productive-to-redundant transition is decodable from hidden states (Qwen3-8B, MATH500): (a) the adjacent phases are linearly separable, and (b) probe scores rise locally at the overthinking boundary rather than drifting gradually. \rev{Position, lexical, and matched-boundary controls are reported in App.~\ref{sec:position_controls}}.}
    \label{fig:motivation}
\end{figure}

We test whether hidden states encode the productive-to-redundant transition. We segment responses into solution attempts, verify their answers, and place the \emph{overthinking boundary} at the end of the first correct solution (FCS; Eq.~\ref{eq:fcs}). Tokens through this boundary are labeled \emph{efficient}; subsequent tokens are labeled \emph{overthinking}. From 61 MATH500 responses, we take the last $20$ efficient and first $20$ overthinking tokens around each boundary and train a logistic-regression probe on Qwen3-8B's Layer-32 hidden states using response-level group cross-validation (CV; details in App.~\ref{sec:position_controls}).

\paragraph{Observations.}
(a) \emph{Phase separation}: the probe reaches 85.9\% accuracy and AUROC 0.928, while t-SNE shows clusters by phase. (b) \emph{Boundary localization}: scores rise at the true boundary but remain flat under permuted alignment (Fig.~\ref{fig:motivation}). (c) \emph{Position controls}: a position-only classifier performs at chance, while position-residualized and position-matched hidden states remain similarly separable. \rev{(d) \emph{Lexical controls}: a 1--2-gram classifier reaches 63.4\% accuracy on the same windows, while the hidden-state probe retains 84.6\% accuracy after lexical features are regressed out. (e) \emph{Matched boundaries}: correct and \emph{incorrect} attempts use similar closing phrases. The probe nevertheless assigns a mean score of 0.80 after FCS boundaries and 0.36 after incorrect solution endings (AUROC 0.89, 95\% CI [0.82, 0.96]). A probe trained to distinguish these boundaries reaches AUROC 0.87 across 725 natural boundaries, compared with 0.54 for a lexical classifier. Within the 11 fail-then-succeed traces, which control for problem and style, it reaches AUROC 0.85.}

\paragraph{Implications.}
\rev{These controls show that late-layer states encode a localized boundary signal that distinguishes correct from incorrect solution endings beyond position and surface wording.} \rev{These findings motivate ROM, which we evaluate end to end on five backbones in Sec.~\ref{sec:experiments}.} The signal can be read online without a verifier, answer extractor, or future tokens.

\subsection{Latent Boundary Control Setup}
\label{sec:problem_setup}

Given a query, a frozen reasoning model $\mathcal{M}$ generates a response $\mathbf{r}=\{r_1,\ldots,r_{T}\}$. Tokens are \emph{efficient} if they contribute to reaching a sufficient solution, and \emph{overthinking} if they follow one and primarily repeat, verify, or explore without improving the final answer. During training, we operationalize this transition through attempt-level correctness labels (Sec.~\ref{sec:labeling}); at inference, ROM predicts it from hidden states alone. Let $\mathbf{h}_t\in\mathbb{R}^{d}$ denote the hidden state of $r_t$ at a fixed backbone layer $L$, and $\mathbf{H}_{1:t}$ the hidden states of the prompt and response through step $t$. ROM learns a streaming detector $p_t = f_\theta(\mathbf{H}_{1:t}) \approx \mathbb{P}(y_t{=}1\mid \mathbf{H}_{1:t})$, where $y_t{=}1$ denotes overthinking.

\subsection{Offline Token-Level Labeling}
\label{sec:labeling}

Following~\citet{chen2025think23overthinkingo1like}, we segment each response into solution attempts $\mathbf{s}=\{s_1,\ldots,s_M\}$ and assign each attempt a correctness label $c_i\in\{0,1\}$ through answer extraction and verification \rev{with GPT-4o (pipeline and prompts in App.~\ref{sec:label_prompts})}. The boundary lies at the end of the first correct solution, whose index is
\begin{equation}\label{eq:fcs}
k^* \;=\; \min\{i \mid c_i=1,\; i\in[1,M]\}.
\end{equation}
Tokens in attempts $\{s_1,\ldots,s_{k^*}\}$ are labeled $0$ (efficient), and tokens in $\{s_{k^*+1},\ldots,s_M\}$ are labeled $1$ (overthinking). We construct boundary supervision from responses containing at least one correct solution (see Sec.~\ref{sec:ablations}). These offline labels identify the reasoning phase: tokens in $s_{k^*}$ remain efficient when they state the correct answer, later tokens are overthinking even if they repeat that answer, and incorrect attempts before $s_{k^*}$ remain efficient. \rev{Sec.~\ref{sec:boundary_validity} evaluates this boundary choice and the reliability of the automatic labels.}

\subsection{Counterfactual Self-Correction: Closing the Attempt-Position Shortcut}
\label{sec:csc}

CSC explicitly separates solution sufficiency from attempt index by introducing wrong$\rightarrow$correct training trajectories. Most original training responses are correct on the first attempt ($k^*{=}1$), making attempt index a shortcut for the boundary label. \rev{We counter this shortcut with two types of training views (Alg.~\ref{alg:augmentation}, App.~\ref{sec:csc_details}). \emph{Counterfactual wrong$\rightarrow$correct views}: we reuse the first attempt if it is incorrect; otherwise, an auxiliary LLM rewrites it into a plausible solution with a different answer. Appending one correct attempt yields an efficient view; appending several yields an overthinking view whose boundary follows the second attempt. \emph{Resampled correct solutions}: single correct attempts form efficient views; reordered subsets form overthinking views with only the first attempt labeled efficient. Sec.~\ref{sec:csc_causal} evaluates CSC's effects on the attempt-index shortcut and natural self-correction.}

\subsection{Streaming Detection}
\label{sec:detection}

At each decoding step, the detector summarizes the prefix of frozen hidden states, updates a memory state, and emits a probability (equations and parameter counts: App.~\ref{sec:head_details}).

\paragraph{Feature extraction.}
\rev{A single-head causal attention projector summarizes the prefix as $\hat{\mathbf{h}}_t=\mathrm{AttnProj}(\mathbf{H}_{1:t})\in\mathbb{R}^{d_p}$, with $d_p{=}1024$. The current state $\mathbf{h}_t$ is projected to a query, while the prefix states, including the prompt, are projected to keys and values.} \rev{The key--value cache requires 4\,KiB per token, and each step performs one attention read over the prefix. The per-step attention cost is therefore linear in prefix length; Sec.~\ref{sec:latency} reports deployment costs.}

\paragraph{Temporal modeling.}
Per-token feed-forward probes can be noisy near the transition to overthinking. To integrate evidence over time, we feed $\hat{\mathbf{h}}_t$ into \rev{a gated recurrent cell with a continuous-time correction}, $\mathbf{m}_t=g(\mathbf{m}_{t-1},\hat{\mathbf{h}}_t)$. \rev{The update follows the closed-form continuous-time formulation~\citep{hasani2022closed}} used in the streaming guardrail of~\citet{li2025kelpstreamingsafeguardlarge}. We initialize $\mathbf{m}_0$ from attention-pooled prompt features, making the detector query-aware from the first generated token.

\paragraph{Classification head and training.}
A linear head maps the memory state to $p_t=\sigma(\mathbf{w}^\top \mathbf{m}_t + b)$. We train the detector with binary cross-entropy over response tokens while keeping the backbone frozen. \rev{Depending on backbone width, the detector contains 19.1--23.1M parameters, equivalent to 0.07--0.25\% of the backbone's parameters (App.~Table~\ref{tab:head_params})}.

\subsection{Boundary-Aware Intervention}
\label{sec:intervention}

At test time, ROM triggers at $t^*=\min\{t \mid p_t > 0.5\}$, the first token whose score crosses the classifier's decision threshold. Because $t^*$ may fall within a sentence or equation, ROM backtraces to the nearest well-formed reasoning boundary $\tilde{t}^*$ (a newline or sentence boundary), discards the tokens decoded after it, and appends a fixed final-answer cue. The model then generates a brief conclusion. \rev{Sec.~\ref{sec:cue_control} isolates the effects of the cue and cut position, while App.~\ref{sec:accounting} details the accounting for discarded tokens}.

\section{Experiments}
\label{sec:experiments}

\subsection{Experimental Setup}
\label{sec:exp_setup}

\paragraph{Backbones and training data.}
We evaluate ROM on five backbones spanning three families and 8B--32B parameters: RL post-trained Qwen3-8B and Qwen3-14B~\citep{yang2025qwen3technicalreport}, Gemma-4-12B-it~\citep{gemmateam2026gemma4}, and the distilled DeepSeek-R1-Distill-Qwen-32B (DS-R1-32B) and DeepSeek-R1-Distill-Llama-8B (DS-Llama-8B)~\citep{deepseek_r1}. \rev{Each backbone has a separately trained head.} We teacher-force the \emph{same} labeled MATH500 traces, generated by QwQ~\citep{qwq32b}, through each backbone to extract hidden states at $83$--$90\%$ of its depth (App.~\ref{sec:impl_details}). \rev{Of 284 training responses, 273 are correct on the first attempt. CSC expands 568 unaugmented views into 740 efficient and 793 overthinking samples.} Only MATH500 contributes labels; the other four benchmarks measure zero-shot task transfer. MATH500 uses a held-out evaluation split.

\paragraph{Benchmarks and protocol.}
We evaluate on MATH500~\citep{hendrycksmath2021} (100 held-out problems), GSM8K~\citep{cobbe2021gsm8k} (full test set, 1{,}319), AIME25~\citep{mathai_aime25} (all 30), GPQA-Diamond~\citep{rein2023gpqa} (full set, 198), and MMLU-Pro~\citep{wang2024mmlu} (70-question validation split\rev{; App.~\ref{sec:scope_app} evaluates 980 test questions}). All methods use identical vLLM decoding settings, $n{=}3$ samples per problem ($n{=}10$ on AIME25), and a uniform $8{,}192$-token output budget. We report accuracy and mean output length with $95\%$ confidence-interval half-widths from a problem-level clustered bootstrap. \rev{For paired comparisons, we bootstrap per-problem differences and apply Holm correction~\citep{holm1979simple} to $p$-values across the 25 settings (App.~\ref{sec:full_results}).} ROM uses a threshold of $0.5$ and boundary-aware backtracing.

\paragraph{Baselines.}
We compare against ten recent early-exit and overthinking-mitigation baselines, covering every inference-time family in App.~Table~\ref{tab:three_properties}: EAT~\citep{wang2025entropylangletextttthinkrangle}, RCPD~\citep{wei2026evolutionthoughttrackingllm}, RPDM~\citep{guan2026rpdm}, NEAT~\citep{liu2026neat}, TERMINATOR~\citep{nagle2026terminatorlearningoptimalexit}, REFRAIN~\citep{sun2025refrain}, Dynasor~\citep{fu2025efficientlyscalingllmreasoning}, PMA~\citep{yan2026pma}, PUMA-RD~\citep{min2026stop}, and RP~\citep{zhang2025reasoningmodelsknowtheyre}. We use official code, released heads or probes, and reference configurations where available (coverage and deviations in App.~\ref{sec:baseline_details}). Truncation-based baselines receive ROM's final-answer cue at their own stopping points. Length columns count completion tokens; probe, trial, \rev{and discarded} tokens are accounted for separately in App.~\ref{sec:accounting}. Code and configurations are available publicly.\footnote{\url{https://github.com/SaFo-Lab/ROM}}

\begin{table*}[t]
\centering
\footnotesize
\renewcommand{\arraystretch}{0.90}
\setlength{\tabcolsep}{4.5pt}
\resizebox{\textwidth}{!}{%
\begin{tabular}{@{}lcccccc@{}}
\toprule
& \multicolumn{3}{c}{\textsc{Math}} & \multicolumn{2}{c}{\textsc{Science}} & \\
\cmidrule(lr){2-4}\cmidrule(lr){5-6}
\textbf{Method} & \textbf{MATH500} & \textbf{GSM8K} & \textbf{AIME25} & \textbf{GPQA-D} & \textbf{MMLU-Pro} & \textbf{Overall} \\
\midrule

\multicolumn{7}{@{}l}{\textbf{Qwen3-8B}} \\
\cmidrule(lr){1-7}
Vanilla & 90.3\ci{4.2} / 4{,}569 & 98.9\ci{0.6} / 1{,}969 & 27.0\ci{14.3} / 6{,}258 & 37.9\ci{5.8} / 5{,}791 & 76.7\ci{8.6} / 2{,}784 & 66.15 / 4{,}274 \\
EAT~\citep{wang2025entropylangletextttthinkrangle} & 83.3\ci{7.0} / \textbf{2{,}088} & 96.3\ci{0.8} / 1{,}563 & 26.0\ci{13.2} / \textbf{3{,}874} & 38.9\ci{5.3} / 4{,}712 & 76.7\ci{8.8} / 2{,}052 & 64.23 / 2{,}858 \\
RCPD~\citep{wei2026evolutionthoughttrackingllm} & 90.3\ci{4.2} / 4{,}586 & 95.0\ci{0.9} / 918 & 27.0\ci{14.5} / 6{,}265 & 36.7\ci{5.7} / 5{,}900 & 75.7\ci{8.8} / 2{,}647 & 64.95 / 4{,}063 \\
RPDM~\citep{guan2026rpdm} & 90.0\ci{4.2} / 4{,}519 & 98.8\ci{0.6} / 1{,}943 & 27.0\ci{14.2} / 6{,}149 & 37.5\ci{5.8} / 6{,}137 & 76.2\ci{8.6} / 2{,}751 & 65.91 / 4{,}300 \\
NEAT~\citep{liu2026neat} & 85.7\ci{6.2} / 4{,}143 & 99.1\ci{0.5} / 2{,}059 & 27.0\ci{15.4} / 7{,}369 & 36.2\ci{5.7} / 6{,}225 & 75.7\ci{9.3} / 2{,}761 & 64.73 / 4{,}511 \\
TERMINATOR~\citep{nagle2026terminatorlearningoptimalexit} & 88.3\ci{5.5} / 3{,}086 & 96.7\ci{0.8} / 710 & 26.3\ci{13.1} / 7{,}203 & 36.7\ci{5.7} / 5{,}210 & 73.8\ci{9.3} / 1{,}991 & 64.37 / 3{,}640 \\
REFRAIN~\citep{sun2025refrain} & 86.7\ci{5.8} / 3{,}345 & 96.7\ci{0.8} / 1{,}576 & 26.7\ci{13.3} / 5{,}997 & 43.3\ci{5.8} / 4{,}643 & 77.6\ci{9.1} / 1{,}749 & 66.18 / 3{,}462 \\
Dynasor~\citep{fu2025efficientlyscalingllmreasoning} & 88.7\ci{4.3} / 3{,}982 & 99.6\ci{0.3} / 1{,}657 & 26.7\ci{14.2} / 6{,}184 & 41.8\ci{5.9} / 5{,}060 & \textbf{80.0\ci{8.7}} / 2{,}094 & 67.33 / 3{,}795 \\
PMA~\citep{yan2026pma} & 84.3\ci{6.2} / 3{,}164 & 99.2\ci{0.4} / 1{,}299 & 25.0\ci{13.3} / 5{,}342 & 38.2\ci{5.8} / 4{,}731 & 74.8\ci{8.6} / 1{,}733 & 64.30 / 3{,}254 \\
PUMA-RD~\citep{min2026stop} & 77.3\ci{6.2} / 3{,}210 & 92.5\ci{1.1} / 1{,}548 & 26.7\ci{14.3} / 6{,}189 & 33.3\ci{5.5} / 5{,}248 & 70.5\ci{8.3} / 2{,}011 & 60.06 / 3{,}641 \\
\rom{} (ours, w/o CSC) & 90.3\ci{4.0} / 2{,}703 & 99.6\ci{0.3} / 669 & 27.0\ci{14.0} / 4{,}587 & 43.4\ci{5.7} / 4{,}285 & 78.1\ci{8.8} / 1{,}095 & 67.70 / 2{,}668 \\
\rowcolor{gray!20} \romcsc{} (ours) & \textbf{90.7\ci{5.0}} / 2{,}412 & \textbf{99.8\ci{0.3}} / \textbf{627} & \textbf{27.3\ci{14.2}} / 4{,}126 & \textbf{44.6\ci{6.0}} / \textbf{3{,}908} & 79.0\ci{8.8} / \textbf{1{,}009} & \textbf{68.29} / \textbf{2{,}416} \\
\midrule
\multicolumn{7}{@{}l}{\textbf{Gemma-4-12B}} \\
\cmidrule(lr){1-7}
Vanilla & 78.3\ci{6.7} / 3{,}789 & 92.5\ci{1.1} / 2{,}095 & 32.3\ci{16.2} / 7{,}636 & 31.3\ci{5.7} / 7{,}005 & 64.3\ci{8.8} / 3{,}787 & 59.75 / 4{,}862 \\
EAT~\citep{wang2025entropylangletextttthinkrangle} & 77.3\ci{6.5} / \textbf{1{,}445} & 95.8\ci{0.9} / 1{,}537 & 32.3\ci{16.2} / 7{,}426 & 32.8\ci{5.2} / 5{,}699 & 70.5\ci{8.3} / 2{,}206 & 61.76 / 3{,}663 \\
RPDM~\citep{guan2026rpdm} & 82.7\ci{5.8} / 3{,}763 & 92.4\ci{1.1} / 2{,}081 & 32.3\ci{16.3} / 7{,}319 & 31.0\ci{5.5} / 6{,}954 & 65.2\ci{8.6} / 3{,}735 & 60.73 / 4{,}770 \\
Dynasor~\citep{fu2025efficientlyscalingllmreasoning} & 83.3\ci{6.2} / 1{,}733 & 96.7\ci{0.8} / 812 & 32.7\ci{13.1} / 6{,}196 & 35.9\ci{6.0} / 4{,}408 & 73.3\ci{8.9} / 1{,}790 & 64.37 / 2{,}988 \\
PMA~\citep{yan2026pma} & 84.0\ci{6.2} / 3{,}392 & 97.5\ci{0.6} / 1{,}199 & 32.3\ci{16.2} / 7{,}341 & 30.1\ci{5.3} / 5{,}060 & 69.0\ci{8.7} / 1{,}949 & 62.60 / 3{,}788 \\
PUMA-RD~\citep{min2026stop} & 76.7\ci{6.8} / 3{,}505 & 88.3\ci{1.3} / 1{,}513 & 32.0\ci{15.4} / 7{,}613 & 28.8\ci{5.2} / 5{,}523 & 66.2\ci{8.9} / 2{,}112 & 58.39 / 4{,}053 \\
\rom{} (ours, w/o CSC) & 78.7\ci{7.5} / 1{,}632 & 92.5\ci{1.2} / 504 & 32.7\ci{15.2} / 5{,}153 & 35.4\ci{5.8} / 4{,}543 & 72.4\ci{8.8} / 1{,}708 & 62.32 / 2{,}708 \\
\rowcolor{gray!20} \romcsc{} (ours) & \textbf{85.0\ci{6.0}} / 1{,}532 & \textbf{98.0\ci{0.7}} / \textbf{479} & \textbf{33.0\ci{14.2}} / \textbf{4{,}358} & \textbf{36.4\ci{5.7}} / \textbf{4{,}090} & \textbf{73.8\ci{8.4}} / \textbf{1{,}633} & \textbf{65.22} / \textbf{2{,}418} \\
\midrule
\multicolumn{7}{@{}l}{\textbf{DS-R1-Distill-Qwen-32B}} \\
\cmidrule(lr){1-7}
Vanilla & 83.3\ci{6.2} / 3{,}367 & 92.8\ci{1.2} / 456 & \textbf{29.7\ci{14.0}} / 6{,}145 & 55.6\ci{6.1} / 3{,}551 & 66.2\ci{8.1} / 1{,}313 & 65.50 / 2{,}966 \\
EAT~\citep{wang2025entropylangletextttthinkrangle} & 77.0\ci{7.0} / \textbf{1{,}693} & 92.5\ci{1.2} / 446 & 27.0\ci{12.3} / \textbf{3{,}762} & 53.9\ci{5.4} / 2{,}950 & 63.8\ci{7.9} / 1{,}030 & 62.83 / 1{,}976 \\
RCPD~\citep{wei2026evolutionthoughttrackingllm} & 83.3\ci{6.2} / 3{,}377 & 93.2\ci{1.2} / 459 & 28.3\ci{13.7} / 6{,}149 & 54.7\ci{5.6} / 3{,}596 & 64.8\ci{7.9} / 1{,}324 & 64.88 / 2{,}981 \\
RPDM~\citep{guan2026rpdm} & 83.3\ci{6.5} / 3{,}344 & 92.8\ci{1.2} / 461 & 28.7\ci{13.5} / 6{,}156 & 55.2\ci{6.0} / 3{,}498 & 66.2\ci{8.1} / 1{,}332 & 65.24 / 2{,}958 \\
Dynasor~\citep{fu2025efficientlyscalingllmreasoning} & \textbf{84.0\ci{6.2}} / 3{,}003 & 92.5\ci{1.3} / 337 & 28.7\ci{13.3} / 5{,}882 & 59.6\ci{6.1} / 2{,}660 & \textbf{72.4\ci{9.1}} / 898 & 67.43 / 2{,}556 \\
PMA~\citep{yan2026pma} & 79.0\ci{6.5} / 2{,}574 & 92.7\ci{1.2} / 457 & 27.7\ci{13.2} / 5{,}174 & 51.5\ci{5.6} / 3{,}098 & 59.5\ci{8.3} / 1{,}135 & 62.08 / 2{,}488 \\
PUMA-RD~\citep{min2026stop} & 81.3\ci{6.5} / 2{,}790 & 92.7\ci{1.2} / 453 & 28.3\ci{13.7} / 6{,}073 & 54.0\ci{5.8} / 3{,}541 & 66.2\ci{7.9} / 1{,}288 & 64.51 / 2{,}829 \\
RP$^{\dagger}$~\citep{zhang2025reasoningmodelsknowtheyre} & 80.7\ci{6.8} / 2{,}440 & 92.2\ci{1.3} / 442 & 28.7\ci{13.8} / 5{,}838 & 57.1\ci{5.7} / 3{,}199 & 70.5\ci{8.4} / 1{,}166 & 65.83 / 2{,}617 \\
\rom{} (ours, w/o CSC) & 83.3\ci{6.3} / 1{,}994 & 93.0\ci{1.2} / 318 & 28.7\ci{12.7} / 4{,}390 & 58.8\ci{5.9} / 1{,}679 & 69.5\ci{8.9} / 636 & 66.66 / 1{,}803 \\
\rowcolor{gray!20} \romcsc{} (ours) & 83.7\ci{6.8} / 1{,}820 & \textbf{93.5\ci{1.2}} / \textbf{300} & 29.0\ci{11.8} / 4{,}015 & \textbf{60.9\ci{5.7}} / \textbf{1{,}590} & 71.9\ci{8.5} / \textbf{592} & \textbf{67.80} / \textbf{1{,}663} \\
\bottomrule
\end{tabular}
}
\caption{Main results: accuracy (\%, $\pm$95\% CI) / mean output tokens, with identical decoding, an 8{,}192-token budget, and $n{=}3$ samples per problem ($n{=}10$ on AIME25). \textbf{Overall}: macro-average across five benchmarks. \textbf{Bold}: highest accuracy / shortest output per model--benchmark setting. \rom{} (w/o CSC) is our ablation. $^{\dagger}$RP: official probes only for DS-R1-32B; coverage details in App.~\ref{sec:baseline_details}. Table~\ref{tab:main_results_full} includes the other two backbones and length CIs; text counts cover all 25 settings.}
\label{tab:main_results}
\end{table*}

\revnextfloat
\begin{figure*}[t]
\centering
\includegraphics[width=\textwidth]{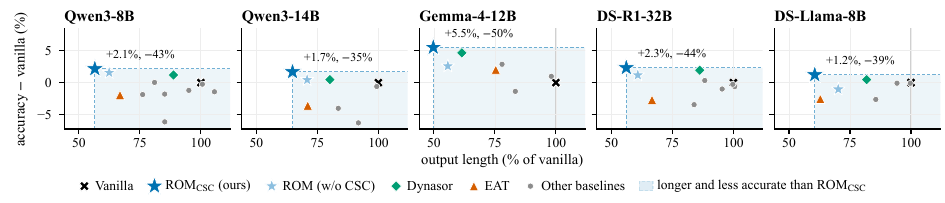}
\caption{Macro-average performance across five benchmarks relative to vanilla decoding (\emph{Overall} in Tables~\ref{tab:main_results} and~\ref{tab:main_results_full}). Each panel shows one backbone and each marker one method. All other methods lie in the shaded region, with longer, less accurate outputs than \romcsc{}. Labels show \romcsc{}'s accuracy gain and length reduction; intervals appear in Table~\ref{tab:paired}.}
\label{fig:macro_pareto}
\end{figure*}

\subsection{Main Results}
\label{sec:backbone_comparison}

\textbf{\romcsc{} reduces output length by 28--77\% (mean 45\%) \rev{while matching or exceeding the accuracy of vanilla decoding in 22 of 25 settings}} (Table~\ref{tab:main_results}). \rev{Paired tests confirm significant length reductions in all 25 settings (intervals in App.~\ref{sec:stats_app}).} In those 22 settings, accuracy increases by $+0.33$ to $+9.52$ points. Macro-average accuracy across the five benchmarks increases on every backbone ($+1.21$ to $+5.47$; Fig.~\ref{fig:macro_pareto}).
\label{sec:analysis}
The benefit scales with the backbone's redundancy: the most verbose backbone under vanilla decoding (Gemma-4-12B) achieves an accuracy gain $4.5\times$ that of the least verbose (DS-Llama-8B; App.~\ref{sec:extra_obs}). ROM also complements RL length control: applied to L1-Qwen3-8B-Max~\citep{aggarwal2025l1controllinglongreasoning}, the same head removes an additional 21.6\% of tokens on MATH500 with unchanged accuracy (Table~\ref{tab:beyond_protocol}).

\subsection{Comparison with Baselines}
\label{sec:baseline_pareto}

\revnextfloat
\begin{figure*}[t]
\centering
\includegraphics[width=\textwidth]{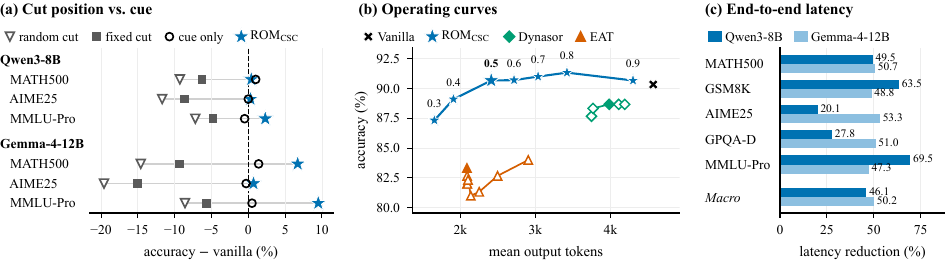}
\caption{Three analyses of the main result. \textbf{(a)}~Accuracy relative to vanilla for the cue-only control, \romcsc{}, and content-blind cuts with matched reasoning retention and the same cue (Sec.~\ref{sec:cue_control}). \textbf{(b)}~Accuracy versus length across parameter sweeps on Qwen3-8B/MATH500; numbers denote ROM thresholds $\tau$ (Sec.~\ref{sec:matched_budget}). \textbf{(c)}~Latency reduction relative to vanilla (Sec.~\ref{sec:latency}).}
\label{fig:controls}
\end{figure*}

\paragraph{Accuracy--length front.}
\romcsc{} lies on the empirical accuracy--length Pareto front in all 25 settings (Table~\ref{tab:main_results}). Averaged across the five benchmarks, \romcsc{} achieves both the highest accuracy and the shortest output on each of the five backbones (Fig.~\ref{fig:macro_pareto}). \rev{To assess sensitivity to sampling variation, we recompute front membership in each bootstrap replicate. The resulting probability that \romcsc{} lies on the accuracy--length front is 0.94--1.00 across the 25 settings (mean 0.992; App.~\ref{sec:stats_app}).}

\paragraph{\revhead{Operating curves.}}
\label{sec:matched_budget}
\rev{At the default settings (Table~\ref{tab:main_results}), Dynasor is the strongest baseline, within $\pm$2.9 accuracy points of \romcsc{} but 43\% longer on average. We also sweep ROM's threshold, Dynasor's stability threshold and chunk size, and EAT's variance threshold on Qwen3-8B/MATH500 (Fig.~\ref{fig:controls}b; App.~\ref{sec:sweep_app}). Every Dynasor point is dominated by a ROM point. At matched length, ROM exceeds Dynasor by 2.1--3.4 accuracy points and EAT by 6.9--8.8 points.}

\paragraph{\revhead{Trigger and early-stop rates.}}
\label{sec:trigger_rates}
\rev{A \emph{trigger} occurs when the stopping rule fires; an \emph{early stop} is a trigger before 95\% of the corresponding sample's vanilla reasoning length. On Qwen3-8B/MATH500, Dynasor triggers on 83.3\% of samples but stops early on only 25.7\%, as its rule waits for a stable, extractable answer. The corresponding rates for \romcsc{} are 89.7\% and 82.7\% (App.~\ref{sec:trigger_app}).}

\paragraph{Confidence and sample diversity.}
\label{sec:entropy_comparison}
\label{sec:passk}
The learned boundary captures information beyond confidence: among 49 MMLU-Pro samples that EAT answers incorrectly on Qwen3-8B, \romcsc{} corrects 6, covering both entropy failure modes (App.~\ref{sec:case_eat_rom}). \rev{To assess sample diversity after early exit, we draw $n{=}16$ samples per problem in six settings (App.~Table~\ref{tab:passk}). \romcsc{}'s pass@16 matches or exceeds vanilla's in every setting, with mean gains of $+3.0$ points on avg@16 and $+1.9$ on pass@16.}

\subsection{\revhead{Cut Position or Answer Cue?}}
\label{sec:cue_control}

\begin{table}[t]
\centering
\footnotesize
\setlength{\tabcolsep}{2.5pt}
\begin{tabular}{@{}lccrr@{}}
\toprule
 & \multicolumn{2}{c}{\textbf{Accuracy (\%)}} & \multicolumn{2}{c}{\textbf{Output tokens}} \\
\cmidrule(lr){2-3}\cmidrule(lr){4-5}
\textbf{Setting} & Base & \romcsc{} & Base & \romcsc{} \\
\midrule
MMLU-Pro, open-ended & 80.21 & \textbf{81.77} & 2{,}457 & \textbf{1{,}587} \\
L1-Max, MATH500      & 90.83 & 90.83          & 2{,}684 & \textbf{2{,}105} \\
\bottomrule
\end{tabular}
\caption{Protocol extensions (Qwen3-8B, $n{=}3$). \emph{Base}: vanilla decoding for open-ended MMLU-Pro (64 non-numerical problems, options removed, GPT-4o judge); L1-Qwen3-8B-Max for MATH500 (40 problems).}
\label{tab:beyond_protocol}
\end{table}

\rev{ROM combines truncation with a final-answer cue. To isolate the effect of cut position, we evaluate three controls on Qwen3-8B and Gemma-4-12B using identical backtracing, cue, continuation, and grading procedures (Fig.~\ref{fig:controls}a; App.~Table~\ref{tab:cue_control}). The \emph{cue-only} control appends the cue to the full vanilla reasoning trace. The \emph{fixed} and \emph{random} controls cut without regard to content while matching ROM's reasoning retention in each setting.}

\rev{The cue alone changes accuracy by $-0.5$ to $+1.4$ points, with changes of at least one point only on MATH500. At matched retention, ROM exceeds vanilla accuracy by 0.3--9.5 points, whereas content-blind cuts reduce accuracy by 4.8--15.0 points (fixed) and 7.2--19.6 points (random). On Gemma-4-12B, where ROM's gains are largest ($+6.7$ on MATH500, $+9.5$ on MMLU-Pro), the cue alone yields gains of only $+1.4$ and $+0.5$ points.} An open-ended evaluation provides consistent evidence: after removing the MMLU-Pro options and grading free-form answers with GPT-4o, \romcsc{} reduces Qwen3-8B's response length by 35.4\% and raises accuracy from 80.21\% to 81.77\% (Table~\ref{tab:beyond_protocol}). \rev{These results show that cut position is critical to preserving accuracy; truncation and answer elicitation alone do not explain ROM's gains.}

\subsection{Ablations}
\label{sec:ablations}

\revnextfloat
\begin{table}[t]
\centering
\footnotesize
\setlength{\tabcolsep}{3.5pt}
\begin{tabular}{@{}lccc@{}}
\toprule
 & \textbf{Distinct} & \textbf{Counter-} & \textbf{Macro length} \\
\textbf{Training set} & \textbf{samples} & \textbf{factual} & \textbf{vs.\ \romcsc{}} \\
\midrule
CSC (full)          & 1{,}533 & \cmark & reference \\
CSC, downsampled   & 568     & \cmark & $+3.2\%$ \\
Resampling only     & 1{,}336 & \xmark & $+19.4\%$ \\
\bottomrule
\end{tabular}
\caption{Size-matched CSC controls (Qwen3-8B; macro-average output length relative to \romcsc{}). The downsampled set matches the unaugmented set's size; \emph{Resampling only} omits the counterfactual branch.}
\label{tab:csc_variants}
\end{table}

\begin{table}[t]
\centering
\footnotesize
\setlength{\tabcolsep}{4pt}
\begin{tabular}{@{}llcc@{}}
\toprule
\multicolumn{2}{@{}l}{\textbf{Configuration}} & \textbf{Acc (\%)} & \textbf{SL} \\
\midrule
\multicolumn{2}{@{}l}{Vanilla (no cut)} & 90.3 & 4569 \\
\rowcolor{gray!20}\multicolumn{2}{@{}l}{\romcsc{} (L32, $\tau{=}0.5$, backtracing)} & \textbf{90.7} & \textbf{2412} \\
\midrule
\emph{Detector} & linear head            & 90.0 & 4105 \\
                & conf.\ MA ($0.98$)     & 70.0 & 1105 \\
                & conf.\ MA ($0.995$)    & 67.7 & \phantom{0}944 \\
\emph{Control}  & w/o backtracing        & 89.7 & 2573 \\
\emph{Layer}    & L22                    & 89.3 & 2108 \\
                & L34                    & 90.7 & 2315 \\
\emph{Threshold}& $0.4$                  & 89.1 & 1909 \\
                & $0.6$                  & 90.7 & 2716 \\
                & $0.7$                  & 91.0 & 3036 \\
\bottomrule
\end{tabular}
\caption{Ablations on MATH500 (Qwen3-8B, 100 held-out problems, $n{=}3$). SL: mean output tokens. Each row below the shaded default changes one component of it. L: detector layer; $\tau$: decision threshold.}
\label{tab:ablations}
\end{table}

\paragraph{\revhead{Validity of the FCS boundary.}}
\label{sec:boundary_validity}
\rev{We test the FCS definition and automatic labels with three checks (App.~\ref{sec:boundary_details}). (i) A verifier-free boundary at the earliest attempt after which the answer remains unchanged matches FCS on 279 of 284 training responses (98.2\%). (ii) In a blind audit of 100 responses, human and pipeline FCS indices agree exactly on 90 and within one solution on 99, with disagreements in both directions and 94.5\% per-solution correctness agreement. (iii) Retraining the Qwen3-8B head with FCS+1, which treats the first post-FCS verification as productive, increases output length by 36.6\% (95\% CI [30.8, 43.4]) for a macro-average accuracy gain of $+0.25$ points over \romcsc{}. This later boundary adds token cost with little accuracy benefit.}

\paragraph{Contributions of CSC.}
\label{sec:csc_causal}
Across the 25 settings, CSC improves accuracy over \rom{} without CSC by $+0.15$ to $+6.33$ points (mean $+1.62$) and further shortens output by 4.4--15.4\%. On DS-Llama-8B, CSC raises macro-average accuracy by 2.25 points to 58.73\%, exceeding vanilla's 57.52\%.

\rev{We isolate trajectory content from data size and class balance through three Qwen3-8B analyses (App.~\ref{sec:csc_app}). \emph{(i) Training-set controls.} CSC retains its length benefit at the unaugmented set's size but loses it without counterfactual trajectories (Table~\ref{tab:csc_variants}). \emph{(ii) Attempt-index shortcut.} After regressing out token position, scores predict ``attempt index $\geq 2$'' with AUROC 0.78 without CSC and 0.53 with CSC, close to chance.}

\rev{\emph{(iii) Natural self-correction.} On 15 natural wrong$\rightarrow$correct traces, truncation before the end of the first correct solution falls from 66.7\% (\rom{} w/o CSC) to 13.3\% (\romcsc{}). The rate barely changes on 130 control traces without self-correction (10.0\% vs.\ 10.8\%; difference-in-differences $-54.1$ points, 95\% CI [$-81$, $-27$]). CSC reduces premature truncation while retaining triggers on 13 of 15 traces. Across eight settings containing 75 problems that vanilla never solves, \romcsc{} triggers less often on these problems than on the remaining problems in seven settings.}

\paragraph{Detector and control.}
On Qwen3-8B/MATH500 (Table~\ref{tab:ablations}), replacing the recurrent cell with a linear classifier over the same attention features reduces token-level training accuracy from 96.1\% to 62.5\%, and the head then rarely triggers. Thresholding a moving average of token confidence instead truncates too aggressively, reducing accuracy to 70.0\%. These results support integrating boundary evidence over time. Removing backtracing reduces accuracy and increases length. Across the probed layers and thresholds of 0.4--0.7, accuracy remains within 1.2 points of vanilla while length reductions range from 34\% to 58\%.

\subsection{\revhead{Efficiency in Streaming Deployment}}
\label{sec:latency}

\rev{We measure end-to-end latency against vanilla on Qwen3-8B (full KV cache) and Gemma-4-12B (sliding-window cache), running the detector in the decoding loop and restoring the retained prefix at each trigger (Fig.~\ref{fig:controls}c; App.~\ref{sec:efficiency_app}).}

\paragraph{\revhead{Latency.}}
\rev{ROM reduces latency by 20--70\% across all ten settings. Including backtracing and regeneration, the reductions are within 2 percentage points of those predicted from decoded-token counts.}

\paragraph{\revhead{Detector cost.}}
\rev{On a real Qwen3-8B AIME25 trace, the head's per-step cost rises from 0.108\,ms at a 1K-token prefix to 0.162\,ms at 16K. As backbone cost also grows, the head remains at 0.30--0.42\% of each step (0.16--0.24\% on Gemma). Both attention costs scale linearly with prefix length. On Qwen3-8B, the head caches 4\,KiB per token versus the backbone's 144\,KiB, a ratio of about 2.8\%.}

\paragraph{\revhead{Memory.}}
\rev{The head adds about 38\,MiB of weights and 4\,KiB of cache per token. Shorter generation offsets this overhead on Qwen3-8B, lowering peak memory by 113\,MiB. On Gemma, peak memory rises by 546\,MiB, mainly from re-prefilling its sliding-window cache after a trigger; the head's cache stays below 22\,MiB.}

\section{Conclusion}
\label{sec:conclusion}

ROM detects the productive-to-redundant transition in a frozen model's hidden states and intervenes at a well-formed boundary. \rev{The signal persists under position and lexical controls.} CSC augments training with wrong$\rightarrow$correct trajectories to preserve useful self-correction. \rev{Separately trained heads share MATH500 supervision and transfer to four unseen benchmarks.} Across five backbones, \romcsc{} reduces response length by 28--77\% \rev{while matching or exceeding vanilla accuracy in 22 of 25 settings. End-to-end latency falls by 20--70\% on two backbones. Matched-length controls show that preserving accuracy depends on where reasoning is cut}.

\section*{Limitations}

\paragraph{Absence of Clear Data Scaling Behavior.}
Our method does not exhibit a pronounced scaling law with respect to training data size: performance on 50\% of the training set is comparable to that on the full dataset. On the other hand, it also implies that our method does not consistently benefit from larger training sets (at least in our dataset scale).

\paragraph{Dependence on Labeling Model Quality.}
Our data generation pipeline relies on a (relatively) high-capability model for automatic solution labeling. In our experiments, we initially explored open-source alternatives such as Llama-3.3-70B-Instruct~\citep{grattafiori2024llama3herdmodels} for labeling, but found its labeling accuracy to be noticeably lower than that of GPT-4o~\citep{openai2024gpt4technicalreport}, which we ultimately adopted. We note that this dependence on a strong labeling model is a common limitation shared by virtually all methods that employ automatic labeling for training data construction.

\section*{Ethics Statement}

By reducing unnecessary reasoning in LRMs, this work lowers computational cost, latency, and energy use, making advanced reasoning more accessible in resource-constrained deployments. Premature stopping can interrupt useful self-correction. CSC trains the detector to preserve wrong$\rightarrow$correct reasoning, while adjustable thresholds control intervention timing. Pre-deployment evaluation guides threshold selection.

All experiments use publicly available benchmarks (MATH500, GSM8K, AIME25, GPQA-Diamond, MMLU-Pro) and models (Qwen3-8B, Qwen3-14B, Gemma-4-12B-it, QwQ-32B, DeepSeek-R1-Distill-Qwen-32B, DeepSeek-R1-Distill-Llama-8B, L1-Qwen3-8B-Max, Llama-3.3-70B-Instruct) in accordance with their published licenses and terms of use. We use official baseline code and published assets where available (App.~\ref{sec:baseline_details}) and access GPT-4o through its API for solution segmentation, correctness verification, counterfactual rewriting, and open-ended evaluation. The datasets contain mathematical and academic questions and, to our knowledge, no personally identifying or offensive content. \rev{The authors conducted the manual audit of boundary labels (App.~\ref{sec:boundary_details});} no external annotators or human subjects were involved, and no new data were collected from people. Licensing details are provided in App.~\ref{sec:impl_details}.

We used AI assistants for coding support and language editing. All scientific claims, experimental designs, and interpretations of results are the authors' own. The authors verified all reported numbers against the experiment logs.

\bibliography{custom}
\bibliographystyle{plainnat}

\appendix
\clearpage

\section{Controls for the Latent Footprint Study}
\label{sec:position_controls}

\begin{figure}[htbp]
    \centering
    \includegraphics[width=\linewidth]{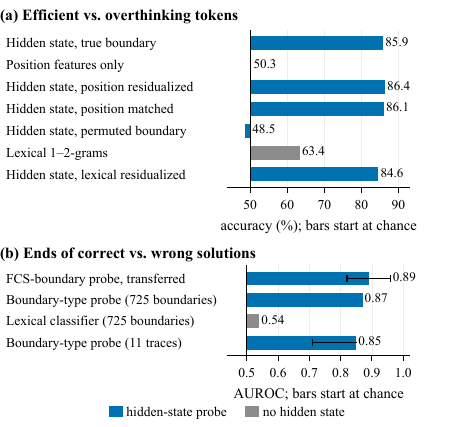}
    \caption{Probes and controls for the diagnostic study in Sec.~\ref{sec:insight} (Qwen3-8B, Layer~32, response-level group CV), complementing Fig.~\ref{fig:motivation}. (a) Hidden-state probes retain 84.6--86.4\% accuracy under position and lexical controls. Position-only features and permuted boundaries yield chance performance, while lexical 1--2-grams reach 63.4\%. (b) Hidden states distinguish correct from incorrect solution endings, whereas a lexical classifier performs near chance. Whiskers show 95\% CIs, except for the 11-trace probe, which uses a one-sided lower bound.}
    \label{fig:insight}
\end{figure}

This appendix details the diagnostic study in Sec.~\ref{sec:insight}. \rev{All experiments use Qwen3-8B's Layer~32 hidden states extracted from teacher-forced training traces. We take a window of $k{=}20$ tokens on each side of the boundary, apply L2 normalization to the hidden states, and train a logistic-regression probe.} We use response-level 5-fold group cross-validation, holding out entire responses so that no token from a test response appears in training. \rev{The sample is drawn from the first of five shards in the labeled training pool (App.~\ref{sec:impl_details}). Of the 62 responses with a labeled efficient-to-overthinking transition, 61 contain at least 20 tokens on each side and qualify for the study ($1{,}220$ tokens per class).} \rev{Fig.~\ref{fig:insight} summarizes the results.}

\paragraph{Position controls.}
The position-only classifier uses three scalar features: absolute token index, normalized token index, and response length. The transition location is excluded because it defines the offline labels and is unavailable at inference. For the position-residual control, we fit a ridge regression within each training fold to remove these features from the hidden states, apply the fitted regression to the held-out fold, and train the classifier on the residuals. For the position-matched control, we group tokens into bins by normalized position and subsample equal numbers of efficient and overthinking tokens within each bin. For the permuted-boundary control, we assign each response another response's transition location as a pseudo-boundary, preserving the local before/after format while breaking alignment with the true transition. The position-only baseline performs near chance ($50.3\%$), whereas the probe retains its accuracy on position-residualized ($86.4\%$) and position-matched ($86.1\%$) hidden states relative to the original states ($85.9\%$; Fig.~\ref{fig:insight}). Permuting the boundaries reduces group-CV accuracy to $48.5\%$.

\revnextfloat

\paragraph{\revhead{Lexical controls.}}
\rev{FCS boundaries coincide with solution-ending discourse (``Therefore, the answer is \ldots''). To isolate the contribution of this wording, we complement the position controls with two lexical controls on the same tokens and folds. The lexical baseline uses logistic regression on 1--2-gram token features. For the lexical-residual control, we regress these features out of the hidden states within each training fold and train the probe on the residuals, following the position-residual procedure. Whereas boundary permutation disrupts both wording and semantics, these controls directly assess the contribution of surface wording. Fig.~\ref{fig:insight}a shows that wording is informative but accounts for little of the hidden-state signal.}

\paragraph{\revhead{Matched solution boundaries.}}
\rev{We further compare boundaries with similar discourse: both correct and incorrect attempts end by stating an answer, but only the former establish correctness. We take the 20 tokens following each boundary from natural, unaugmented traces in the training pool (App.~\ref{sec:csc_details}) and conduct three complementary analyses. \emph{(i) Transfer.} We apply the probe from Sec.~\ref{sec:insight}, trained only on FCS boundaries, to both boundary types and compare the mean overthinking scores. \emph{(ii) Direct probe.} We train a probe to distinguish the two boundary types under response-level group CV, using all non-final attempt endings in the 284-response pool (29 incorrect, 696 correct). A lexical classifier uses the same windows. \emph{(iii) Within-trace comparison.} In the 11 fail-then-succeed traces, incorrect and correct solution endings share the same response, controlling for problem, difficulty, and style. On these traces, the direct probe's AUROC has a one-sided 95\% lower bound of 0.71. Observation~(e) in Sec.~\ref{sec:insight} reports the results, and Fig.~\ref{fig:insight}b plots the AUROCs.}

\section{Implementation, Training, and Reproducibility Details}
\label{sec:impl_details}

\paragraph{Software stack.} ROM is implemented in PyTorch~2.9 with Transformers~4.57. Token-level labels are derived from QwQ~\citep{qwq32b} outputs generated with vLLM under an $8{,}192$-token output budget. For each of the five target backbones, we tokenize the same labeled traces with its tokenizer, map segment-level labels to the resulting tokens, and replay the traces through the frozen backbone under teacher forcing to extract aligned hidden states. This design combines QwQ's rich multi-solution trajectories with the representations of the backbone that each detector will monitor. At evaluation time, the head processes that backbone's own generated traces. We precompute and cache $\{\mathbf{h}_t\}$ for every training sample on each backbone, eliminating repeated backbone forward passes during detector training.

\subsection{\revhead{Offline Labeling Pipeline and Prompts}}
\label{sec:label_prompts}

\rev{Labels are produced once, offline, with GPT-4o (Azure OpenAI deployment \texttt{gpt-4o}, model version \texttt{gpt-4o-2024-11-20}, API version \texttt{2024-12-01-preview}) at temperature $0$.}

\paragraph{\revhead{Segmentation.}} \rev{We follow the three-stage procedure of~\citet{chen2025think23overthinkingo1like}. \emph{(1)~Split.} The response is divided into numbered steps, and the model groups consecutive steps into solutions that end with explicit answers. A verification or alternative derivation that reaches an answer counts as a separate solution, even if the answer is unchanged. \emph{(2)~Step judge.} The model checks whether each proposed final step explicitly states an answer. \emph{(3)~Split judge.} The model checks whether each proposed boundary separates two reasoning processes and whether the preceding part is logically complete. The boxes below reproduce excerpts of the core instructions in their original wording, with line breaks added and omissions marked [...]. The in-context demonstrations used at each stage are released with the code.}

\transcriptcaption{\rev{Stage~1 (split) instructions.}}
\begin{transcriptbox}{\rev{Segmentation prompt, stage 1 (excerpt)}}
\begin{scriptsize}\revon
\begin{verbatim}
Here is a student's response to a problem, please read
the response and insert some xml tags into the response
to separate different solutions.

Definition of "solution":
A "solution" is a complete process of solving the
problem, and a solution should end with an answer
explicitly. [...] If there is no clear answer from the
analysis, then the answer should be "none".

Notes to extract answer:
1. Don't care about the correctness of the solution,
   just extract them and present the answer in it.
2. The student may verify or reflect the solution after
   getting one answer, or try another way to solve the
   problem. Once you see a valid answer, mark it as a
   solution, and leave the verification, reflection
   and different exploration to the next solution.
   Note: When a reflection process arrives at an
   answer, whether or not that answer is consistent
   with the previous answer, the process *should be*
   regarded as a solution, too.
3. Different solution may contain the same answer,
   just extract them all.
[...]
Rules to extract answer:
1. Whenever you see an explicit answer, that's a
   solution.
2. Double-Check or verification is considered a
   solution.
3. Some steps at the ending of the response may not
   contain any reasoning process and just a conclusion
   of the current solution, these steps should be
   merged to the last soluition.
4. Single Step Solution: if one single step contains a
   complete reasoning (or double-check) process and an
   answer, then it should be treated as a single step
   soluition.
\end{verbatim}
\end{scriptsize}
\end{transcriptbox}

\transcriptcaption{\rev{Stage~2 (step judge) and stage~3 (split judge) instructions.}}
\begin{transcriptbox}{\rev{Segmentation prompts, stages 2 and 3 (excerpts)}}
\begin{scriptsize}\revon
\begin{verbatim}
[Stage 2] Here is a single step of a student's response
to a problem, please read the step and judge whether
the step contains the expected answer explicitly.
Rules to judge:
1. Do not consider the correctness of the expected
   answer.
2. Only steps that explicitly include a conclusion are
   considered to contain an answer.
3. \\boxed is a strong sign of containing an answer.

[Stage 3] Here is a small snippet of a student's
solution to a problem, and there is an ending mark in
the snippet. Please read the step and judge whether
the ending mark is reasonable.
Rules to define a reasonable ending mark:
1. A reasonable ending mark should separate two
   thinking process.
2. The ending of the part before the ending mark should
   must be logically complete and contain an answer
   explicitly.
3. If the part after the ending mark is a simple
   conclusion, e.g. just presenting the final answer,
   then the ending mark is not reasonable.
\end{verbatim}
\end{scriptsize}
\end{transcriptbox}

\paragraph{\revhead{Per-attempt correctness.}} \rev{For each solution, we read the answer from its \texttt{\textbackslash boxed\{\}} expression when available; otherwise, GPT-4o extracts it from the end of the solution. We compare the extracted answer with the reference through a fixed cascade. Exact string matches are accepted as correct, purely numeric answers are compared using a rule-based mathematical-equivalence check, and solutions without an extractable answer are marked incorrect. For all other answers, the rule-based check determines correctness when it succeeds; otherwise, GPT-4o judges equivalence. App.~\ref{sec:boundary_details} validates the resulting labels against human annotations.}

\subsection{\revhead{Training Pool and CSC}}
\label{sec:csc_details}

\paragraph{\revhead{Training pool.}} \rev{We divide MATH500 into five shards of 100 problems and split each shard 80/20, yielding 400 training problems and 100 held-out evaluation problems. QwQ generates one response per training problem, and the pipeline above segments and judges each response. We exclude 106 single-attempt responses, which have no continuation to label, and 10 responses without a correct attempt, for which no FCS boundary can be defined. The resulting pool contains 284 responses and 1{,}009 attempts, with $k^*{=}1$ in 273 responses and $k^*{\geq}2$ in 11. Without augmentation, each response yields two views: an efficient view truncated at the FCS, with all tokens labeled $0$, and an overthinking view containing the full response. These 568 samples form the training set for \rom{} (w/o CSC).}

\revnextfloat
\begin{algorithm}[htbp]
\small
\caption{Counterfactual Self-Correction (CSC), as implemented. $\mathrm{Lab}(\cdot)$ applies the FCS rule from Eq.~\ref{eq:fcs} within each view: tokens up to and including the first correct attempt are labeled $0$, and all subsequent tokens are labeled $1$.}
\label{alg:augmentation}
\begin{algorithmic}[1]
\REQUIRE Responses with attempts $\mathbf{s}=(s_1,\ldots,s_M)$, $M\geq 2$, and correctness labels $\mathbf{c}=(c_1,\ldots,c_M)$; rewriting LLM $\mathcal{R}$
\ENSURE Efficient set $\mathcal{D}_{\text{eff}}$ (all tokens labeled $0$); overthinking set $\mathcal{D}_{\text{over}}$
\STATE $\mathcal{D}_{\text{eff}}\gets \emptyset$;\; $\mathcal{D}_{\text{over}}\gets \emptyset$
\FOR{each response $(\mathbf{s},\mathbf{c})$}
\STATE $C \gets (s_i : c_i=1)$ \COMMENT{correct attempts, in order of appearance}
\IF{$C=\emptyset$}
  \STATE \textbf{continue} \COMMENT{no FCS boundary can be defined}
\ENDIF
\STATE $k^* \gets \min\{i \mid c_i=1\}$ \COMMENT{FCS index}
\STATE \COMMENT{\textit{Wrong attempt $w$: natural if the response starts with one, counterfactual otherwise}}
\IF{$c_1=0$}
  \STATE $w \gets s_1$
\ELSE
  \STATE $w \gets \mathcal{R}(s_1)$: rewrite $s_1$ so that its final answer changes (same problem, style, and format; up to 10 tries); $w\gets\varnothing$ if the answer never changes
\ENDIF
\STATE \COMMENT{\textit{Resampled correct solutions}}
\STATE Add $(s_{k^*})$ and $(u)$ with $u\sim C$ to $\mathcal{D}_{\text{eff}}$
\STATE Add the original response $\mathbf{s}$ to $\mathcal{D}_{\text{over}}$ with labels $\mathrm{Lab}(\mathbf{s})$
\IF{$|C|\geq 2$}
  \STATE Draw $U$: $j\sim\mathrm{Unif}\{2,\ldots,|C|\}$ attempts sampled from $C$ without replacement, in random order; add $U$ to $\mathcal{D}_{\text{over}}$ with labels $\mathrm{Lab}(U)$
\ENDIF
\STATE \COMMENT{\textit{Counterfactual wrong$\rightarrow$correct views}}
\IF{$w\neq\varnothing$}
  \STATE Add $(w,u)$ with $u\sim C$ to $\mathcal{D}_{\text{eff}}$ \COMMENT{self-correction through the first correct solution: all tokens $0$}
  \IF{$|C|\geq 2$}
    \STATE Draw $U$ as above; add $(w,U)$ to $\mathcal{D}_{\text{over}}$ with labels $\mathrm{Lab}((w,U))$ \COMMENT{$w$ and the first element of $U$ are $0$, the rest $1$}
  \ENDIF
\ELSE
  \STATE If $|C|\geq 2$: add one more single-solution view to $\mathcal{D}_{\text{eff}}$ and one more resampled view $U$ to $\mathcal{D}_{\text{over}}$
\ENDIF
\STATE In every assembled view, replace the first sentence of each relocated attempt with a transition phrase and remove the final-answer block from attempts that are no longer last
\ENDFOR
\STATE Remove exact duplicates;\; \textbf{return} $\mathcal{D}_{\text{eff}},\mathcal{D}_{\text{over}}$
\end{algorithmic}
\end{algorithm}

\paragraph{\revhead{CSC.}} \rev{Alg.~\ref{alg:augmentation} implements the two view types introduced in Sec.~\ref{sec:csc}, referred to here as the \emph{counterfactual} and \emph{resampling} components. Applied to the 284-response pool, it yields 740 efficient samples (477 single correct solutions and 263 wrong$\rightarrow$correct pairs) and 793 overthinking samples, totaling 1{,}533 samples. The original response is assigned to $\mathcal{D}_{\text{over}}$; when it ends at its first correct solution, all of its tokens retain the efficient label.} The released CSC training data are available at \url{https://huggingface.co/datasets/xinyan-wang/ROM}.

\paragraph{\revhead{Counterfactual rewriting.}} \rev{The rewriting model $\mathcal{R}$ uses the same GPT-4o deployment, with temperature $1.0$, top-$p$ $1.0$, and a maximum of $4{,}096$ output tokens. A rewrite is accepted when the model returns an answer and reports that it differs from the original. We allow up to 10 attempts; if none is accepted, we omit the counterfactual attempt for that response. The prompt is reproduced below.}

\transcriptcaption{\rev{Prompt used to synthesize the counterfactual wrong attempt.}}
\begin{transcriptbox}{\rev{CSC rewriting prompt (original wording, line breaks added)}}
\begin{scriptsize}\revon
\begin{verbatim}
You are given a mathematical solution:
{original_solution}.
- Make context-consistent edits to this solution so
  that the reasoning still looks natural, coherent, and
  aligned with the original problem/context, but the
  **final conclusion/answer** becomes different from
  the current answer.
- Keep the original style, tone, structure, and overall
  approach as much as possible.
- Do not mention anything or compare with the original
  solution.
- Output the modified solution, final answer, and
  whether it is equal to the original one, format:
  {"solution":MODIFIED SOLUTION, "answer":ANSWER,
   "equal": true/false}
Output:
\end{verbatim}
\end{scriptsize}
\end{transcriptbox}

\paragraph{\revhead{Transitions.}} \rev{To connect reassembled attempts, we replace the first sentence of each relocated attempt with a phrase sampled from one of three fixed lists of 20 phrases each. A self-check phrase introduces a correct attempt after an incorrect one (e.g., ``Wait, let me check if there are any mistakes.''); an alternative-method phrase introduces the original first attempt when moved later (e.g., ``Alternatively, maybe there's another way to approach this problem.''); and an opening phrase introduces a later attempt moved to the front (e.g., ``Alright, let's tackle this problem.''). All analyses of the detector's learned signal (App.~\ref{sec:position_controls}, App.~\ref{sec:csc_app}) use natural, unaugmented traces, independently of these templated transitions.}

\subsection{\revhead{Detection Head: Equations and Parameter Counts}}
\label{sec:head_details}

\rev{Let $\mathbf{u}_1,\ldots,\mathbf{u}_{P+t}$ denote the layer-$L$ hidden states of the $P$ prompt tokens and the first $t$ response tokens. Thus, $\mathbf{h}_t=\mathbf{u}_{P+t}$. We set the projection dimension to $d_p{=}1024$.}

\paragraph{\revhead{Attention projector.}} \rev{We use a single causal attention head with no additional positional terms:}
\begingroup\revon
\begin{align}
\mathbf{q}_i &= W_q\mathbf{u}_i+\mathbf{b}_q,\quad
\mathbf{k}_i = W_k\mathbf{u}_i+\mathbf{b}_k, \nonumber\\
\mathbf{v}_i &= W_v\mathbf{u}_i+\mathbf{b}_v, \nonumber\\
\alpha_{ij} &= \frac{\exp(\mathbf{q}_i^\top\mathbf{k}_j/\sqrt{d_p})}{\sum_{j'\leq i}\exp(\mathbf{q}_i^\top\mathbf{k}_{j'}/\sqrt{d_p})},\quad j\leq i, \nonumber\\
\hat{\mathbf{u}}_i &= \textstyle\sum_{j\leq i}\alpha_{ij}\mathbf{v}_j,
\qquad \hat{\mathbf{h}}_t=\hat{\mathbf{u}}_{P+t},
\end{align}
\endgroup
\rev{Here, $W_q,W_k,W_v\in\mathbb{R}^{d_p\times d}$. At each decoding step $t$, we compute $\mathbf{q},\mathbf{k},\mathbf{v}$ for the new token and reuse the cached keys and values of all preceding tokens.}

\paragraph{\revhead{Initial state.}} \rev{The memory is initialized from attention-pooled prompt features:}
\begingroup\revon
\begin{align}
\beta_j &= \frac{\exp(\mathbf{w}_s^\top\hat{\mathbf{u}}_j)}{\sum_{j'\leq P}\exp(\mathbf{w}_s^\top\hat{\mathbf{u}}_{j'})}, \nonumber\\
\mathbf{m}_0 &= \tanh\!\Big(W_p\textstyle\sum_{j\leq P}\beta_j\hat{\mathbf{u}}_j+\mathbf{b}_p\Big).
\end{align}
\endgroup

\paragraph{\revhead{Recurrent update and discretization.}} \rev{The memory update uses the logistic function $\sigma$ and elementwise product $\odot$:}
\begingroup\revon
\begin{align}
\mathbf{z}_t &= \sigma(W_z\hat{\mathbf{h}}_t+U_z\mathbf{m}_{t-1}+\mathbf{b}_z), \nonumber\\
\mathbf{g}_t &= \sigma(W_g\hat{\mathbf{h}}_t+U_g\mathbf{m}_{t-1}+\mathbf{b}_g), \nonumber\\
\tilde{\mathbf{m}}_t &= \tanh(W_h\hat{\mathbf{h}}_t+U_h(\mathbf{g}_t\odot\mathbf{m}_{t-1})+\mathbf{b}_h), \nonumber\\
\bar{\mathbf{m}}_t &= (1-\mathbf{z}_t)\odot\mathbf{m}_{t-1}+\mathbf{z}_t\odot\tilde{\mathbf{m}}_t, \nonumber\\
\mathbf{m}_t &= \bar{\mathbf{m}}_t+\Delta_t\,(\bar{\mathbf{m}}_t-\mathbf{m}_{t-1}).
\end{align}
\endgroup
\rev{The first four lines define the gated update, with all matrices in $\mathbb{R}^{d_p\times d_p}$. The final line adds a discretized continuous-time correction, taking an explicit step of size $\Delta_t$ along the update direction. We map each response to the unit time interval, setting $\Delta_1{=}0$ and using $\Delta_t{=}1/(T-1)$ for complete training traces with $T$ response tokens. During streaming deployment, we fix $\Delta_t{=}1/(T_{\max}-1)$ using the response budget $T_{\max}{=}8{,}192$.}

\paragraph{\revhead{Output.}} \rev{The head outputs $p_t=\mathrm{softmax}(W_o\mathbf{m}_t+\mathbf{b}_o)_1$, with $W_o\in\mathbb{R}^{2\times d_p}$. This two-class softmax is equivalent to the sigmoid form in Sec.~\ref{sec:detection}, with $\mathbf{w}$ given by the difference between the two rows of $W_o$.}

\paragraph{\revhead{Parameter count.}} \rev{The detector contains $3(d\,d_p+d_p)$ parameters in the projector, $6d_p^2+3d_p$ in the recurrent cell, $d_p^2+2d_p$ in the prompt initializer, and $2d_p+2$ in the output layer, for a total of $3\,d\,d_p+7d_p^2+10d_p+2$. Table~\ref{tab:head_params} lists the parameter counts and monitored layer for each backbone. A fixed depth rule selects layers at $83$--$90\%$ of backbone depth, without per-backbone tuning. The weights occupy 37--44\,MiB in \texttt{bfloat16} (37\,MiB on Gemma-4-12B).}

\revnextfloat
\begin{table}[t]
\centering
\footnotesize
\setlength{\tabcolsep}{3.5pt}
\begin{tabular}{@{}lrrrrr@{}}
\toprule
\textbf{Backbone} & $d$ & \textbf{Layers} & $L$ & \textbf{Head} & \textbf{Share} \\
\midrule
Qwen3-8B     & 4096 & 36 & 32 & 19.93M & 0.24\% \\
Qwen3-14B    & 5120 & 40 & 36 & 23.08M & 0.16\% \\
Gemma-4-12B  & 3840 & 48 & 40 & 19.15M & 0.16\% \\
DS-R1-32B    & 5120 & 64 & 56 & 23.08M & 0.07\% \\
DS-Llama-8B  & 4096 & 32 & 28 & 19.93M & 0.25\% \\
\bottomrule
\end{tabular}
\caption{Detection-head size per backbone. $d$: backbone hidden size; $L$: layer whose hidden states the head reads; \emph{Head}: trainable parameters; \emph{Share}: head parameters relative to backbone parameters.}
\label{tab:head_params}
\end{table}

\subsection{Training and Decoding}

\paragraph{Training hyperparameters.} We train each head for 20 epochs on the 1{,}533 CSC samples (App.~\ref{sec:csc_details}) using AdamW with a learning rate of $5\!\times\!10^{-5}$, weight decay $0.1$, $(\beta_1,\beta_2){=}(0.9,0.95)$, $\epsilon{=}10^{-8}$, and a cosine schedule with a warmup ratio of $0.1$. The per-device batch size is $8$, with $4$ gradient-accumulation steps (effective batch size $32$). We clip gradients to a maximum norm of $1.0$, use \texttt{bfloat16} mixed precision with the backbone frozen, cap sequences at $8{,}192$ tokens, and set the seed to $46$. With cached hidden states, training completes in under one hour on a single NVIDIA A100 (80\,GB) GPU.

\paragraph{Evaluation decoding.} All backbones are served with vLLM at temperature $0.6$, top-$p$ $0.95$, top-$k$ $20$, and seed $46$. These settings are shared by every method in Tables~\ref{tab:main_results} and~\ref{tab:main_results_full}---vanilla, all ten baselines, and both ROM variants---to isolate differences in stopping behavior. \rev{TERMINATOR and NEAT retain their official two-segment runtimes, which apply the output budget to the reasoning segment (App.~\ref{sec:extra_obs}).}

\paragraph{Inference-time intervention.} After truncating at the boundary $\tilde{t}^*$, we append a fixed final-answer cue (\texttt{\textbackslash n\allowbreak</think>\textbackslash n\allowbreak---\textbackslash n\allowbreak\textbackslash n\allowbreak\#\#\#\ Final Answer\textbackslash n\allowbreak\textbackslash n\allowbreak\$\$\textbackslash n\allowbreak\textbackslash boxed\{}) to elicit a concise answer block for mathematical and multiple-choice evaluations. For open-ended MMLU-Pro, we append only \texttt{\textbackslash n</think>} and grade the resulting free-form answer with GPT-4o. Gemma-4 uses channel markers in place of \texttt{</think>}, so its cue closes the thought channel with the corresponding family-specific marker. A format profile for each model family determines the answer scaffold. Backtracing starts at the trigger token and rewinds to immediately after the nearest preceding newline. If the preceding sentence is incomplete or begins with a reflective marker (e.g., ``Wait''), backtracing continues one sentence further to the previous sentence terminator.

\subsection{\revhead{Computational Accounting}}
\label{sec:accounting}

For truncation-based methods, the length columns of Tables~\ref{tab:main_results} and~\ref{tab:main_results_full} count completion tokens in the retained prefix, answer cue, and regenerated tail. \rev{These counts measure user-visible output. The accounting below covers the additional computation required to produce it.}

\paragraph{\revhead{Discarded tokens (ROM).}} \rev{ROM discards the tokens between the backtraced cut $\tilde{t}^*$ and the trigger $t^*$, adding $t^*-\tilde{t}^*$ decoded tokens beyond the reported output length. Backtracing rewinds at most to the previous sentence terminator, limiting this overhead to one or two sentences per triggered sample. The mean value of $t^*-\tilde{t}^*$ is 25--45 tokens per triggered sample across benchmarks. Discarded tokens account for 1.6\% (Qwen3-14B) to 4.0\% (DS-Llama-8B) of all decoded tokens, with the largest share on backbones that produce the shortest responses.}

\paragraph{Probe and trial tokens (baselines).} Dynasor, PMA, and PUMA-RD perform additional probe or trial decoding at candidate exit points. These tokens are logged separately and excluded from the length columns, favoring the baselines in output-length comparisons. Dynasor's probe tokens average $24$--$31\%$ of its completion tokens across backbones. EAT's reported lengths include stepwise entropy probing. \rev{ROM's detector, TERMINATOR's token-level head, NEAT's exit neurons, and the rank- and entropy-based rules of RCPD and RPDM require no extra probe tokens.} RP also requires no extra decoding, but runs an online answer-position detector and extractor at every checkpoint.

\paragraph{\revhead{Prefill and replay.}} \rev{For backbones that support cache cropping, ROM crops the KV cache to the retained prefix and prefills only the cue before resuming decoding. For backbones such as Gemma-4 with sliding-window layers, it restores the retained prefix by re-prefilling after a trigger. The latency and memory measurements in Sec.~\ref{sec:latency} include this cost.}

\paragraph{\revhead{Total cost.}} \rev{The end-to-end measurements in Sec.~\ref{sec:latency} and App.~\ref{sec:efficiency_app} cover the complete ROM generation process, including the detector, discarded tokens, cache cropping or re-prefill, cue prefill, and regenerated tail.}

\subsection{Reproducibility and Licensing}
\label{sec:repro}

The complete training and evaluation pipelines, together with the CSC training data, are released in an anonymized package at
\url{https://github.com/SaFo-Lab/ROM}.
The package includes YAML configurations specifying all training and evaluation hyperparameters
(\texttt{configs/train.yaml} and \texttt{configs/eval.yaml}),
the 1{,}533 training samples
(\texttt{data/train\_efficient.jsonl} and
\texttt{data/train\_overthinking.jsonl}),
\rev{the labeling prompts and augmentation script described in this appendix,}
and the exact commands for reproduction, documented in the README
(\texttt{python -m rom.train} and
\texttt{python -m rom.eval}).

Our code is released under the MIT License.
The public benchmarks are used under their original licenses:
MIT for MATH500, GSM8K, and MMLU-Pro;
Apache 2.0 for AIME25; and
CC BY 4.0 for GPQA-Diamond.
Qwen3-8B, Qwen3-14B, QwQ-32B, and
L1-Qwen3-8B-Max are used under the Apache 2.0 License.
DeepSeek-R1-Distill-Qwen-32B and
DeepSeek-R1-Distill-Llama-8B are released under the MIT License;
the former is derived from Qwen2.5-32B under the Apache 2.0
License, while the latter is derived from Llama-3.1-8B-Base
and is additionally subject to the Llama 3.1 Community License.
Llama-3.3-70B-Instruct is used under the Llama 3.3 Community
License.
Gemma-4-12B-it is used under the Gemma Terms of Use, and
GPT-4o is accessed through the Azure OpenAI API under the applicable
terms of use.
Baseline methods additionally use their official code releases
and published assets, as documented in App.~\ref{sec:baseline_details}.

\section{\revhead{Validity of the FCS Boundary}}
\label{sec:boundary_details}

\revnextfloat
\begin{table}[t]
\centering
\footnotesize
\setlength{\tabcolsep}{3pt}
\begin{tabular}{@{}>{\raggedright\arraybackslash}p{0.34\linewidth}>{\raggedright\arraybackslash}p{0.28\linewidth}>{\raggedright\arraybackslash}p{0.31\linewidth}@{}}
\toprule
\textbf{Check} & \textbf{Agreement with FCS} & \textbf{Disagreements} \\
\midrule
Answer-stability boundary (284 responses) & 279/284 (98.2\%) & 4 earlier, 1 later; no stable answer is wrong \\
\addlinespace
Blind human audit (100 responses, 380 solutions) & exact 90/100; within one solution 99/100 & 5 earlier, 5 later; 9 of 10 by one solution \\
\addlinespace
Per-solution correctness, human vs.\ pipeline & 359/380 (94.5\%) & --- \\
\midrule
\textbf{Sensitivity} & \textbf{Output length} & \textbf{Macro accuracy} \\
\midrule
Detector retrained on FCS+1 (Qwen3-8B) & $+36.6\%$ {\scriptsize[$+30.8$, $+43.4$]} & $+0.25$ points \\
\bottomrule
\end{tabular}
\caption{Validation of the FCS boundary (Sec.~\ref{sec:boundary_validity}). Top: agreement of two alternative boundary annotations with the pipeline's FCS index on the training pool before augmentation. Bottom: changes relative to \romcsc{} after shifting the boundary one attempt later and retraining (95\% CI in brackets).}
\label{tab:boundary}
\end{table}

\rev{The FCS rule places the boundary at the end of the first correct solution, identified by an automatic verifier. The three checks in Table~\ref{tab:boundary} support both the reliability of these labels and the effectiveness of this boundary for efficient reasoning. All use the 284-response training pool described in App.~\ref{sec:csc_details}.}

\paragraph{\revhead{Answer-stability boundary.}} \rev{A verifier-free alternative places the boundary at the earliest attempt after which the answer remains unchanged. We extract each attempt's answer and select the first that matches all subsequent answers. The resulting labels agree with FCS on 99.5\% of tokens. The response-level disagreement rate in Table~\ref{tab:boundary} (1.8\%) is below the prespecified 5\% threshold for retraining with this alternative. This agreement emerges despite variation in where correct answers occur: a correct attempt can appear anywhere in a response, and the final attempt is judged incorrect in 3 responses.}

\paragraph{\revhead{Blind human audit.}} \rev{We independently re-annotated a stratified sample of 100 responses, with the pipeline's correctness labels and FCS indices hidden from the annotator. The sample includes all 11 responses with $k^*{\geq}2$, covering the cases that require self-correction, and 89 randomly selected from the remaining 273. For each response, we checked the segmentation, judged each solution against the reference answer, and identified the first correct solution. Reweighting the agreement rates in Table~\ref{tab:boundary} to reflect the pool's composition yields 92.1\% exact agreement and 99.6\% agreement within one solution. The annotator accepted the pipeline's segmentation in 97.3\% of cases.}

\paragraph{\revhead{Shifted boundary (FCS+1).}} \rev{Both checks support the alignment of the FCS boundary. We further test a later boundary by relabeling the first post-FCS attempt, typically a verification, as efficient, rebuilding the CSC set with the same augmentation seed, and retraining the Qwen3-8B head under the original protocol. This shift reduces truncation to 6.7--51.2\% of samples across benchmarks. Table~\ref{tab:boundary} (bottom) reports the changes in output length and accuracy across the five benchmarks. ROM also supports adjusting this accuracy--length trade-off directly through the decision threshold, without retraining (Table~\ref{tab:ablations}).}

\section{Baseline Implementations and Provenance}
\label{sec:baseline_details}

\begin{table}[t]
\centering
\setlength{\tabcolsep}{2.5pt}
\renewcommand{\arraystretch}{1.08}
\scriptsize
\begin{tabular}{@{}>{\raggedright\arraybackslash}p{0.335\linewidth}>{\centering\arraybackslash}p{0.135\linewidth}>{\centering\arraybackslash}p{0.09\linewidth}>{\centering\arraybackslash}p{0.145\linewidth}>{\centering\arraybackslash}p{0.155\linewidth}@{}}
\toprule
\textbf{Method family} & \textbf{Frozen backbone} & \textbf{Token level} & \textbf{Extraction free} & \textbf{Transition supervised} \\
\midrule
\emph{Length retraining:} L1, O1-Pruner & \xmark & \xmark & \cmark & \xmark \\
\emph{Answer-defined triggers:} Dynasor, DEER, PUMA-RD, PMA, RP & \cmark & \xmark & \xmark & \xmark \\
\emph{Surface heuristics (step level):} EAT, REFRAIN & \cmark & \xmark & \cmark & \xmark \\
\emph{Surface heuristics (token level):} SyncThink, RCPD, RPDM, NEAT & \cmark & \cmark & \cmark & \xmark \\
\emph{Answer-arrival head:} TERMINATOR & \cmark & \cmark & \cmark & \xmark \\
\midrule
\emph{Latent-boundary head:} \textbf{ROM (ours)} & \cmark & \cmark & \cmark & \cmark \\
\bottomrule
\end{tabular}
\caption{Positioning against prior overthinking mitigation, grouped by control-signal family; rows merge identical profiles, and all methods are cited in Sec.~\ref{sec:related}. \emph{Extraction free}: no online intermediate-answer extraction or probe/trial decoding. \emph{Transition supervised}: the stopping signal is trained on the productive-to-redundant boundary itself, not on length objectives, decoding statistics, or answer arrival. \cmark/\xmark: satisfies/violates.}
\label{tab:three_properties}
\end{table}

We use official baseline releases where available and specify implementation details and adaptations below. The backbone coverage in Table~\ref{tab:main_results} follows the available heads, neuron sets, probes, and runtime support. TERMINATOR, NEAT, and REFRAIN are evaluated on Qwen3-8B; RP on DS-R1-32B; and RCPD on Qwen3-8B and DS-R1-32B. The remaining methods cover all five backbones.

\paragraph{EAT~\citep{wang2025entropylangletextttthinkrangle}.}
We use training-free entropy stabilization with EMA timescale $\alpha{=}0.2$, variance threshold $\delta{=}0.3$, a minimum exit distance of 20 steps, and answer regeneration capped at 1{,}000 tokens. Reported lengths include the cost of stepwise entropy probing.

\paragraph{TERMINATOR~\citep{nagle2026terminatorlearningoptimalexit}.}
We use the official vLLM plugin server with the released \texttt{Terminator-Qwen3-8B} head and default hyperparameters, including sliding-window majority voting and the built-in exit phrase. Our benchmark client provides an OpenAI-compatible interface to this server. The released head determines the baseline's coverage of Qwen3-8B. Requests are issued sequentially, matching the official server's support for one sequence at a time.

\paragraph{NEAT~\citep{liu2026neat}.}
We retain the official runtime logic and thresholds. To evaluate NEAT on Qwen3-8B, we reconstruct its exit-neuron set following the paper's procedure: MATH-train calibration traces, \texttt{</think>} log-probability gain attribution, and temporal filtering. The released neuron set covers DeepSeek-7B, and the original identification script depends on unreleased files. We port the official V0-engine runtime to a V1 logits processor while preserving its actions and thresholds.

\paragraph{RCPD~\citep{wei2026evolutionthoughttrackingllm}.}
We implement the paper's four termination-token rank rules, for which no official code is available. These rules depend only on the prefix, making detection on teacher-forced traces equivalent to online stopping. We also evaluate RCPD on DS-R1-32B, consistent with its training-free, backbone-agnostic design and the original paper's evaluation of DeepSeek-R1-family models.

\paragraph{REFRAIN~\citep{sun2025refrain}.}
Our implementation follows the paper's blank-line step segmentation, reflection-word and answer-cue gating, MiniLM maximum-cosine step redundancy, and SW-UCB threshold adaptation over $\tau\in[0.60,0.80]$. No official code is available. For the unspecified parameters $\lambda$/$C$/$W$, we use $0.1$/$\sqrt{2}$/$10$.

\paragraph{Dynasor~\citep{fu2025efficientlyscalingllmreasoning}.}
We preserve the official repository's probe suffix, uncertainty word list, answer-stability window, and termination strings, using the official CLI defaults (stability threshold 3, chunk size 64). Probes use the main protocol's sampling parameters with fixed per-probe seeds; the official client leaves seeds unset.

\paragraph{RPDM~\citep{guan2026rpdm}.}
We directly implement the paper's full-vocabulary entropy and local/global ratio RPDI with $W{=}512$ and $\lambda{=}2.0$. No official code or boundary-symbol set is available, so we evaluate the trigger at every token.

\paragraph{PMA~\citep{yan2026pma}.}
We implement the paper's three-phase pipeline of geometry analysis, probing, and continuation, using a DeepSeek-R1-Distill-Qwen-1.5B proxy for answer-entropy probes ($k{=}5$). We use the published default constants where specified and document the remaining choices in the released code.

\paragraph{PUMA-RD~\citep{min2026stop}.}
We run the official six-stage pipeline with its released redundancy-detector checkpoint and two-stage stopping rules. The official configurations cover the DeepSeek-R1-Distill and Qwen3 families. We use \texttt{DS-32B.conf} on its corresponding backbone, DS-R1-32B; \texttt{DS-7B.conf} on DS-Llama-8B; and \texttt{Q30B-T.conf} on Qwen3-8B, Qwen3-14B, and Gemma-4-12B. All three configurations share the published confidence threshold $0.98$, $\epsilon{=}0.03$, minimum stop step $10$, and similarity threshold $\tau_{\text{sim}}{=}0.35$. They differ only in the consecutive-redundancy stop parameter $m$ ($0$ for \texttt{Q30B-T}, $1$ for \texttt{DS-7B}, $4$ for \texttt{DS-32B}). A think-tag adapter handles Gemma's channel markers, and memory-related environment overrides preserve the stopping behavior.

\paragraph{RP~\citep{zhang2025reasoningmodelsknowtheyre}.}
We use the officially released DS-R1-32B probes with the paper's default confidence threshold of $0.85$; a full threshold sweep yields consistent behavior. The trigger operates on extracted intermediate answers and therefore requires an online answer-position detector and extractor, both of which ROM avoids.

\section{Full Main-Table Results, Intervals, and Paired Tests}
\label{sec:full_results}

Table~\ref{tab:main_results_full} reports results for all five backbones, including Qwen3-14B and DS-Llama-8B, which are omitted from Table~\ref{tab:main_results} for space. It provides 95\% confidence intervals for accuracy and mean output length. We compute these intervals with a \emph{problem}-level clustered bootstrap, resampling problems with replacement and keeping all $n$ samples from each problem together ($B{=}20{,}000$ replicates, seed~46). This procedure accounts for correlation among samples of the same problem. We report the $2.5$/$97.5$ percentile interval as a symmetric half-width around the point estimate. Near saturation, this notation can place the nominal upper limit above 100 (e.g., Qwen3-8B \romcsc{} on GSM8K, $99.77{\pm}0.26$); the limit is interpreted as clipped to 100. To preserve the same treatment of within-problem correlation, \rev{every test in this paper resamples problems}. Reported lengths include generations stopped at the $8{,}192$-token cap, which censors their lengths (e.g., Gemma-4-12B vanilla decoding on AIME25, $7{,}636$ mean tokens).

\begin{table*}[p]
\centering
\scriptsize
\setlength{\tabcolsep}{4pt}
\resizebox{\textwidth}{!}{%
\begin{tabular}{@{}lcccccc@{}}
\toprule
& \multicolumn{3}{c}{\textsc{Math}} & \multicolumn{2}{c}{\textsc{Science}} & \\
\cmidrule(lr){2-4}\cmidrule(lr){5-6}
\textbf{Method} & \textbf{MATH500} & \textbf{GSM8K} & \textbf{AIME25} & \textbf{GPQA-D} & \textbf{MMLU-Pro} & \textbf{Overall} \\
\midrule

\multicolumn{7}{@{}l}{\textbf{Qwen3-8B}} \\
\cmidrule(lr){1-7}
Vanilla & 90.3\ci{4.2} / 4{,}569\ci{484} & 98.9\ci{0.6} / 1{,}969\ci{60} & 27.0\ci{14.3} / 6{,}258\ci{550} & 37.9\ci{5.8} / 5{,}791\ci{212} & 76.7\ci{8.6} / 2{,}784\ci{449} & 66.15 / 4{,}274 \\
EAT~\citep{wang2025entropylangletextttthinkrangle} & 83.3\ci{7.0} / \textbf{2{,}088\ci{94}} & 96.3\ci{0.8} / 1{,}563\ci{28} & 26.0\ci{13.2} / \textbf{3{,}874\ci{178}} & 38.9\ci{5.3} / 4{,}712\ci{126} & 76.7\ci{8.8} / 2{,}052\ci{331} & 64.23 / 2{,}858 \\
RCPD~\citep{wei2026evolutionthoughttrackingllm} & 90.3\ci{4.2} / 4{,}586\ci{498} & 95.0\ci{0.9} / 918\ci{50} & 27.0\ci{14.5} / 6{,}265\ci{550} & 36.7\ci{5.7} / 5{,}900\ci{256} & 75.7\ci{8.8} / 2{,}647\ci{456} & 64.95 / 4{,}063 \\
RPDM~\citep{guan2026rpdm} & 90.0\ci{4.2} / 4{,}519\ci{484} & 98.8\ci{0.6} / 1{,}943\ci{57} & 27.0\ci{14.2} / 6{,}149\ci{526} & 37.5\ci{5.8} / 6{,}137\ci{213} & 76.2\ci{8.6} / 2{,}751\ci{428} & 65.91 / 4{,}300 \\
NEAT~\citep{liu2026neat} & 85.7\ci{6.2} / 4{,}143\ci{534} & 99.1\ci{0.5} / 2{,}059\ci{62} & 27.0\ci{15.4} / 7{,}369\ci{395} & 36.2\ci{5.7} / 6{,}225\ci{211} & 75.7\ci{9.3} / 2{,}761\ci{469} & 64.73 / 4{,}511 \\
TERMINATOR~\citep{nagle2026terminatorlearningoptimalexit} & 88.3\ci{5.5} / 3{,}086\ci{589} & 96.7\ci{0.8} / 710\ci{14} & 26.3\ci{13.1} / 7{,}203\ci{427} & 36.7\ci{5.7} / 5{,}210\ci{248} & 73.8\ci{9.3} / 1{,}991\ci{358} & 64.37 / 3{,}640 \\
REFRAIN~\citep{sun2025refrain} & 86.7\ci{5.8} / 3{,}345\ci{429} & 96.7\ci{0.8} / 1{,}576\ci{29} & 26.7\ci{13.3} / 5{,}997\ci{540} & 43.3\ci{5.8} / 4{,}643\ci{183} & 77.6\ci{9.1} / 1{,}749\ci{269} & 66.18 / 3{,}462 \\
Dynasor~\citep{fu2025efficientlyscalingllmreasoning} & 88.7\ci{4.3} / 3{,}982\ci{457} & 99.6\ci{0.3} / 1{,}657\ci{54} & 26.7\ci{14.2} / 6{,}184\ci{561} & 41.8\ci{5.9} / 5{,}060\ci{274} & \textbf{80.0\ci{8.7}} / 2{,}094\ci{414} & 67.33 / 3{,}795 \\
PMA~\citep{yan2026pma} & 84.3\ci{6.2} / 3{,}164\ci{329} & 99.2\ci{0.4} / 1{,}299\ci{32} & 25.0\ci{13.3} / 5{,}342\ci{449} & 38.2\ci{5.8} / 4{,}731\ci{194} & 74.8\ci{8.6} / 1{,}733\ci{221} & 64.30 / 3{,}254 \\
PUMA-RD~\citep{min2026stop} & 77.3\ci{6.2} / 3{,}210\ci{349} & 92.5\ci{1.1} / 1{,}548\ci{22} & 26.7\ci{14.3} / 6{,}189\ci{537} & 33.3\ci{5.5} / 5{,}248\ci{190} & 70.5\ci{8.3} / 2{,}011\ci{212} & 60.06 / 3{,}641 \\
\rom{} (ours, w/o CSC) & 90.3\ci{4.0} / 2{,}703\ci{410} & 99.6\ci{0.3} / 669\ci{41} & 27.0\ci{14.0} / 4{,}587\ci{452} & 43.4\ci{5.7} / 4{,}285\ci{168} & 78.1\ci{8.8} / 1{,}095\ci{384} & 67.70 / 2{,}668 \\
\rowcolor{gray!20} \romcsc{} (ours) & \textbf{90.7\ci{5.0}} / 2{,}412\ci{392} & \textbf{99.8\ci{0.3}} / \textbf{627\ci{26}} & \textbf{27.3\ci{14.2}} / 4{,}126\ci{407} & \textbf{44.6\ci{6.0}} / \textbf{3{,}908\ci{223}} & 79.0\ci{8.8} / \textbf{1{,}009\ci{324}} & \textbf{68.29} / \textbf{2{,}416} \\
\midrule
\multicolumn{7}{@{}l}{\textbf{Qwen3-14B}} \\
\cmidrule(lr){1-7}
Vanilla & \textbf{89.3\ci{5.2}} / 4{,}038\ci{463} & 98.3\ci{0.7} / 1{,}570\ci{52} & \textbf{37.0\ci{15.0}} / 5{,}847\ci{534} & 53.4\ci{6.1} / 6{,}060\ci{226} & 84.3\ci{7.9} / 2{,}362\ci{421} & 72.46 / 3{,}975 \\
EAT~\citep{wang2025entropylangletextttthinkrangle} & 82.7\ci{6.8} / \textbf{2{,}251\ci{122}} & 96.7\ci{0.8} / 1{,}284\ci{24} & 33.7\ci{13.2} / 4{,}483\ci{425} & 49.2\ci{6.1} / \textbf{4{,}166\ci{103}} & 82.4\ci{8.1} / 1{,}937\ci{248} & 68.91 / 2{,}824 \\
RPDM~\citep{guan2026rpdm} & 88.3\ci{5.5} / 4{,}001\ci{459} & 98.3\ci{0.7} / 1{,}585\ci{54} & 36.3\ci{14.8} / 5{,}792\ci{529} & 52.5\ci{6.1} / 6{,}013\ci{231} & 83.8\ci{7.9} / 2{,}349\ci{414} & 71.86 / 3{,}948 \\
Dynasor~\citep{fu2025efficientlyscalingllmreasoning} & 86.7\ci{5.8} / 3{,}129\ci{441} & 98.8\ci{0.6} / 1{,}058\ci{47} & 36.3\ci{14.2} / 5{,}104\ci{516} & 56.7\ci{5.7} / 4{,}903\ci{268} & 86.2\ci{7.6} / 1{,}728\ci{377} & 72.94 / 3{,}184 \\
PMA~\citep{yan2026pma} & 83.7\ci{6.3} / 3{,}288\ci{334} & 97.9\ci{0.7} / 1{,}163\ci{34} & 32.3\ci{13.3} / 4{,}917\ci{452} & 47.5\ci{6.1} / 5{,}382\ci{209} & 81.0\ci{8.3} / 1{,}856\ci{233} & 68.47 / 3{,}321 \\
PUMA-RD~\citep{min2026stop} & 79.3\ci{6.5} / 3{,}374\ci{356} & 94.2\ci{1.0} / 1{,}247\ci{21} & 35.0\ci{15.4} / 5{,}936\ci{548} & 45.0\ci{5.8} / 5{,}694\ci{197} & 77.6\ci{8.6} / 1{,}994\ci{216} & 66.21 / 3{,}649 \\
\rom{} (ours, w/o CSC) & 88.7\ci{4.8} / 2{,}694\ci{408} & 99.2\ci{0.4} / 892\ci{43} & 36.3\ci{15.2} / 4{,}126\ci{441} & 55.0\ci{5.9} / 4{,}718\ci{183} & 85.2\ci{7.6} / 1{,}641\ci{318} & 72.89 / 2{,}814 \\
\rowcolor{gray!20} \romcsc{} (ours) & 89.0\ci{4.3} / 2{,}437\ci{389} & \textbf{99.6\ci{0.3}} / \textbf{810\ci{31}} & 36.7\ci{14.7} / \textbf{3{,}728\ci{398}} & \textbf{58.2\ci{6.0}} / 4{,}382\ci{214} & \textbf{87.1\ci{7.4}} / \textbf{1{,}523\ci{287}} & \textbf{74.13} / \textbf{2{,}576} \\
\midrule
\multicolumn{7}{@{}l}{\textbf{Gemma-4-12B}} \\
\cmidrule(lr){1-7}
Vanilla & 78.3\ci{6.7} / 3{,}789\ci{535} & 92.5\ci{1.1} / 2{,}095\ci{106} & 32.3\ci{16.2} / 7{,}636\ci{277} & 31.3\ci{5.7} / 7{,}005\ci{210} & 64.3\ci{8.8} / 3{,}787\ci{533} & 59.75 / 4{,}862 \\
EAT~\citep{wang2025entropylangletextttthinkrangle} & 77.3\ci{6.5} / \textbf{1{,}445\ci{61}} & 95.8\ci{0.9} / 1{,}537\ci{38} & 32.3\ci{16.2} / 7{,}426\ci{343} & 32.8\ci{5.2} / 5{,}699\ci{180} & 70.5\ci{8.3} / 2{,}206\ci{280} & 61.76 / 3{,}663 \\
RPDM~\citep{guan2026rpdm} & 82.7\ci{5.8} / 3{,}763\ci{526} & 92.4\ci{1.1} / 2{,}081\ci{101} & 32.3\ci{16.3} / 7{,}319\ci{371} & 31.0\ci{5.5} / 6{,}954\ci{218} & 65.2\ci{8.6} / 3{,}735\ci{516} & 60.73 / 4{,}770 \\
Dynasor~\citep{fu2025efficientlyscalingllmreasoning} & 83.3\ci{6.2} / 1{,}733\ci{362} & 96.7\ci{0.8} / 812\ci{44} & 32.7\ci{13.1} / 6{,}196\ci{664} & 35.9\ci{6.0} / 4{,}408\ci{294} & 73.3\ci{8.9} / 1{,}790\ci{406} & 64.37 / 2{,}988 \\
PMA~\citep{yan2026pma} & 84.0\ci{6.2} / 3{,}392\ci{536} & 97.5\ci{0.6} / 1{,}199\ci{28} & 32.3\ci{16.2} / 7{,}341\ci{378} & 30.1\ci{5.3} / 5{,}060\ci{217} & 69.0\ci{8.7} / 1{,}949\ci{319} & 62.60 / 3{,}788 \\
PUMA-RD~\citep{min2026stop} & 76.7\ci{6.8} / 3{,}505\ci{379} & 88.3\ci{1.3} / 1{,}513\ci{48} & 32.0\ci{15.4} / 7{,}613\ci{256} & 28.8\ci{5.2} / 5{,}523\ci{202} & 66.2\ci{8.9} / 2{,}112\ci{269} & 58.39 / 4{,}053 \\
\rom{} (ours, w/o CSC) & 78.7\ci{7.5} / 1{,}632\ci{276} & 92.5\ci{1.2} / 504\ci{8} & 32.7\ci{15.2} / 5{,}153\ci{273} & 35.4\ci{5.8} / 4{,}543\ci{234} & 72.4\ci{8.8} / 1{,}708\ci{243} & 62.32 / 2{,}708 \\
\rowcolor{gray!20} \romcsc{} (ours) & \textbf{85.0\ci{6.0}} / 1{,}532\ci{328} & \textbf{98.0\ci{0.7}} / \textbf{479\ci{39}} & \textbf{33.0\ci{14.2}} / \textbf{4{,}358\ci{749}} & \textbf{36.4\ci{5.7}} / \textbf{4{,}090\ci{288}} & \textbf{73.8\ci{8.4}} / \textbf{1{,}633\ci{328}} & \textbf{65.22} / \textbf{2{,}418} \\
\midrule
\multicolumn{7}{@{}l}{\textbf{DS-R1-Distill-Qwen-32B}} \\
\cmidrule(lr){1-7}
Vanilla & 83.3\ci{6.2} / 3{,}367\ci{472} & 92.8\ci{1.2} / 456\ci{7} & \textbf{29.7\ci{14.0}} / 6{,}145\ci{395} & 55.6\ci{6.1} / 3{,}551\ci{234} & 66.2\ci{8.1} / 1{,}313\ci{304} & 65.50 / 2{,}966 \\
EAT~\citep{wang2025entropylangletextttthinkrangle} & 77.0\ci{7.0} / \textbf{1{,}693\ci{88}} & 92.5\ci{1.2} / 446\ci{6} & 27.0\ci{12.3} / \textbf{3{,}762\ci{152}} & 53.9\ci{5.4} / 2{,}950\ci{132} & 63.8\ci{7.9} / 1{,}030\ci{138} & 62.83 / 1{,}976 \\
RCPD~\citep{wei2026evolutionthoughttrackingllm} & 83.3\ci{6.2} / 3{,}377\ci{469} & 93.2\ci{1.2} / 459\ci{7} & 28.3\ci{13.7} / 6{,}149\ci{395} & 54.7\ci{5.6} / 3{,}596\ci{207} & 64.8\ci{7.9} / 1{,}324\ci{334} & 64.88 / 2{,}981 \\
RPDM~\citep{guan2026rpdm} & 83.3\ci{6.5} / 3{,}344\ci{468} & 92.8\ci{1.2} / 461\ci{8} & 28.7\ci{13.5} / 6{,}156\ci{395} & 55.2\ci{6.0} / 3{,}498\ci{212} & 66.2\ci{8.1} / 1{,}332\ci{341} & 65.24 / 2{,}958 \\
Dynasor~\citep{fu2025efficientlyscalingllmreasoning} & \textbf{84.0\ci{6.2}} / 3{,}003\ci{448} & 92.5\ci{1.3} / 337\ci{5} & 28.7\ci{13.3} / 5{,}882\ci{455} & 59.6\ci{6.1} / 2{,}660\ci{197} & \textbf{72.4\ci{9.1}} / 898\ci{181} & 67.43 / 2{,}556 \\
PMA~\citep{yan2026pma} & 79.0\ci{6.5} / 2{,}574\ci{299} & 92.7\ci{1.2} / 457\ci{7} & 27.7\ci{13.2} / 5{,}174\ci{366} & 51.5\ci{5.6} / 3{,}098\ci{156} & 59.5\ci{8.3} / 1{,}135\ci{217} & 62.08 / 2{,}488 \\
PUMA-RD~\citep{min2026stop} & 81.3\ci{6.5} / 2{,}790\ci{318} & 92.7\ci{1.2} / 453\ci{5} & 28.3\ci{13.7} / 6{,}073\ci{387} & 54.0\ci{5.8} / 3{,}541\ci{218} & 66.2\ci{7.9} / 1{,}288\ci{274} & 64.51 / 2{,}829 \\
RP$^{\dagger}$~\citep{zhang2025reasoningmodelsknowtheyre} & 80.7\ci{6.8} / 2{,}440\ci{469} & 92.2\ci{1.3} / 442\ci{8} & 28.7\ci{13.8} / 5{,}838\ci{486} & 57.1\ci{5.7} / 3{,}199\ci{191} & 70.5\ci{8.4} / 1{,}166\ci{310} & 65.83 / 2{,}617 \\
\rom{} (ours, w/o CSC) & 83.3\ci{6.3} / 1{,}994\ci{330} & 93.0\ci{1.2} / 318\ci{6} & 28.7\ci{12.7} / 4{,}390\ci{440} & 58.8\ci{5.9} / 1{,}679\ci{104} & 69.5\ci{8.9} / 636\ci{128} & 66.66 / 1{,}803 \\
\rowcolor{gray!20} \romcsc{} (ours) & 83.7\ci{6.8} / 1{,}820\ci{259} & \textbf{93.5\ci{1.2}} / \textbf{300\ci{7}} & 29.0\ci{11.8} / 4{,}015\ci{404} & \textbf{60.9\ci{5.7}} / \textbf{1{,}590\ci{109}} & 71.9\ci{8.5} / \textbf{592\ci{134}} & \textbf{67.80} / \textbf{1{,}663} \\
\midrule
\multicolumn{7}{@{}l}{\textbf{DS-R1-Distill-Llama-8B}} \\
\cmidrule(lr){1-7}
Vanilla & 80.7\ci{6.5} / 897\ci{139} & 91.0\ci{1.3} / 245\ci{7} & 17.3\ci{11.7} / 2{,}613\ci{271} & 35.7\ci{5.7} / 2{,}426\ci{186} & 62.9\ci{9.5} / 484\ci{112} & 57.52 / 1{,}333 \\
EAT~\citep{wang2025entropylangletextttthinkrangle} & 77.3\ci{6.4} / 712\ci{52} & 90.5\ci{1.3} / 238\ci{5} & 13.0\ci{9.5} / \textbf{1{,}488\ci{69}} & 33.7\ci{5.1} / \textbf{1{,}360\ci{52}} & 60.5\ci{8.9} / 391\ci{41} & 55.00 / 838 \\
RPDM~\citep{guan2026rpdm} & 80.0\ci{6.7} / 881\ci{137} & 91.1\ci{1.3} / 248\ci{8} & 17.3\ci{11.2} / 2{,}627\ci{273} & 35.4\ci{5.6} / 2{,}398\ci{184} & 63.3\ci{9.3} / 492\ci{117} & 57.42 / 1{,}329 \\
Dynasor~\citep{fu2025efficientlyscalingllmreasoning} & 80.3\ci{6.8} / 754\ci{107} & 90.9\ci{1.4} / 198\ci{6} & 17.7\ci{12.0} / 2{,}288\ci{235} & 37.2\ci{5.9} / 1{,}859\ci{145} & 63.8\ci{9.8} / 356\ci{78} & 57.99 / 1{,}091 \\
PMA~\citep{yan2026pma} & 78.3\ci{6.7} / 771\ci{115} & 90.8\ci{1.3} / 252\ci{7} & 15.3\ci{10.8} / 2{,}107\ci{219} & 32.0\ci{5.3} / 2{,}154\ci{167} & 58.1\ci{9.1} / 419\ci{92} & 54.91 / 1{,}141 \\
PUMA-RD~\citep{min2026stop} & 79.7\ci{6.5} / 796\ci{119} & 91.0\ci{1.3} / 241\ci{7} & 17.3\ci{12.2} / 2{,}461\ci{252} & 35.9\ci{6.0} / 2{,}337\ci{178} & 63.3\ci{9.7} / 448\ci{99} & 57.43 / 1{,}257 \\
\rom{} (ours, w/o CSC) & 78.7\ci{6.4} / 638\ci{98} & 91.5\ci{1.2} / 181\ci{7} & 14.3\ci{10.9} / 1{,}836\ci{193} & 36.0\ci{5.5} / 1{,}694\ci{132} & 61.9\ci{9.2} / 329\ci{75} & 56.48 / 936 \\
\rowcolor{gray!20} \romcsc{} (ours) & \textbf{81.0\ci{6.1}} / \textbf{542\ci{81}} & \textbf{92.2\ci{1.2}} / \textbf{161\ci{6}} & \textbf{18.0\ci{11.7}} / 1{,}607\ci{169} & \textbf{37.7\ci{5.4}} / 1{,}443\ci{113} & \textbf{64.8\ci{8.9}} / \textbf{283\ci{64}} & \textbf{58.73} / \textbf{807} \\
\bottomrule
\end{tabular}
}
\caption{Complete version of Table~\ref{tab:main_results}: all five backbones---including Qwen3-14B and DS-R1-Distill-Llama-8B, which are deferred from the main text---with 95\% CIs for both accuracy and mean output length.}
\label{tab:main_results_full}
\end{table*}

\subsection{\revhead{Paired Tests and Pareto-Front Uncertainty}}
\label{sec:stats_app}

\paragraph{\revhead{Paired differences.}} \rev{The intervals in Table~\ref{tab:main_results_full} describe individual methods. Pairing methods on the same problems accounts for their shared variation in problem difficulty, the largest source of variance, and directly measures uncertainty in their difference. We compute per-problem differences in mean accuracy and output length for each pair of methods, then resample these differences with the problem-level bootstrap ($B{=}20{,}000$, seed~46) to obtain a 95\% percentile interval and a two-sided $p$-value. For each comparison and metric, we apply Holm correction~\citep{holm1979simple} across the 25 settings. Per-backbone macro-averages use a stratified bootstrap that resamples problems within each benchmark.}

\paragraph{\revhead{Results.}} \rev{Table~\ref{tab:paired} reports the paired comparisons in all settings. Relative to vanilla, \romcsc{} achieves significantly higher accuracy in 14 settings, with no significant decrease in any setting. The point estimates in Table~\ref{tab:main_results_full} show accuracy above vanilla in 22 of 25 settings, the highest accuracy in 19, and the shortest output in 16. These descriptive counts complement the paired significance tests.}

\revnextfloat\begin{table*}[t]
\centering
\footnotesize
\setlength{\tabcolsep}{6pt}
\begin{tabular}{@{}llccccc@{}}
\toprule
 & & \multicolumn{3}{c}{\textbf{Accuracy: \romcsc{} $-$ vanilla (points)}} & \multicolumn{2}{c}{\textbf{Output length change (\%)}} \\
\cmidrule(lr){3-5}\cmidrule(l){6-7}
\textbf{Backbone} & \textbf{Benchmark} & $\Delta$ & 95\% CI & $p_{\text{Holm}}$ & $\Delta$ & 95\% CI \\
\midrule
Qwen3-8B & MATH500 & $+0.34$ & [$-1.67$, $+2.67$] & 1.00 & $-47.2$ & [$-52.1$, $-42.1$] \\
 & GSM8K & $+0.91$ & [$+0.38$, $+1.49$] & 0.006 & $-68.2$ & [$-68.9$, $-67.4$] \\
 & AIME25 & $+0.33$ & [$-1.00$, $+1.67$] & 1.00 & $-34.1$ & [$-37.9$, $-30.1$] \\
 & GPQA-D & $+6.73$ & [$+4.55$, $+9.09$] & $<$0.003 & $-32.5$ & [$-34.9$, $-30.1$] \\
 & MMLU-Pro & $+2.38$ & [$-0.95$, $+6.67$] & 1.00 & $-63.8$ & [$-69.7$, $-57.7$] \\
 & \emph{Macro} & $+2.14$ & [$+1.13$, $+3.22$] & $<$0.001 & $-43.5$ & [$-45.3$, $-41.6$] \\
\midrule
Qwen3-14B & MATH500 & $-0.33$ & [$-2.33$, $+1.67$] & 1.00 & $-39.6$ & [$-45.1$, $-33.8$] \\
 & GSM8K & $+1.24$ & [$+0.66$, $+1.87$] & $<$0.003 & $-48.4$ & [$-49.6$, $-47.2$] \\
 & AIME25 & $-0.33$ & [$-2.33$, $+1.33$] & 1.00 & $-36.2$ & [$-40.3$, $-32.1$] \\
 & GPQA-D & $+4.88$ & [$+2.53$, $+7.41$] & $<$0.003 & $-27.7$ & [$-29.8$, $-25.5$] \\
 & MMLU-Pro & $+2.85$ & [$+0.95$, $+5.24$] & 0.045 & $-35.5$ & [$-42.3$, $-27.9$] \\
 & \emph{Macro} & $+1.67$ & [$+0.83$, $+2.52$] & $<$0.001 & $-35.2$ & [$-37.2$, $-33.2$] \\
\midrule
Gemma-4-12B & MATH500 & $+6.67$ & [$+3.33$, $+10.33$] & $<$0.003 & $-59.6$ & [$-64.1$, $-54.8$] \\
 & GSM8K & $+5.46$ & [$+4.55$, $+6.42$] & $<$0.003 & $-77.1$ & [$-78.2$, $-76.0$] \\
 & AIME25 & $+0.67$ & [$-1.00$, $+2.67$] & 1.00 & $-42.9$ & [$-50.7$, $-34.6$] \\
 & GPQA-D & $+5.05$ & [$+3.03$, $+7.07$] & $<$0.003 & $-41.6$ & [$-44.4$, $-38.7$] \\
 & MMLU-Pro & $+9.52$ & [$+4.76$, $+14.76$] & $<$0.003 & $-56.9$ & [$-61.6$, $-51.9$] \\
 & \emph{Macro} & $+5.47$ & [$+4.19$, $+6.83$] & $<$0.001 & $-50.3$ & [$-53.1$, $-47.2$] \\
\midrule
DS-R1-32B & MATH500 & $+0.34$ & [$-1.67$, $+2.67$] & 1.00 & $-45.9$ & [$-50.5$, $-41.0$] \\
 & GSM8K & $+0.73$ & [$+0.40$, $+1.11$] & $<$0.003 & $-34.2$ & [$-35.2$, $-33.3$] \\
 & AIME25 & $-0.67$ & [$-3.33$, $+1.67$] & 1.00 & $-34.7$ & [$-38.6$, $-30.7$] \\
 & GPQA-D & $+5.38$ & [$+3.20$, $+7.74$] & $<$0.003 & $-55.2$ & [$-57.1$, $-53.3$] \\
 & MMLU-Pro & $+5.71$ & [$+2.38$, $+9.52$] & 0.013 & $-54.9$ & [$-60.4$, $-48.8$] \\
 & \emph{Macro} & $+2.30$ & [$+1.23$, $+3.41$] & $<$0.001 & $-43.9$ & [$-46.1$, $-41.6$] \\
\midrule
DS-Llama-8B & MATH500 & $+0.33$ & [$-1.00$, $+1.67$] & 1.00 & $-39.6$ & [$-44.9$, $-33.6$] \\
 & GSM8K & $+1.16$ & [$+0.73$, $+1.64$] & $<$0.003 & $-34.3$ & [$-35.7$, $-32.8$] \\
 & AIME25 & $+0.67$ & [$-0.67$, $+2.00$] & 1.00 & $-38.5$ & [$-42.3$, $-34.4$] \\
 & GPQA-D & $+2.02$ & [$+0.84$, $+3.20$] & 0.006 & $-40.5$ & [$-43.3$, $-37.6$] \\
 & MMLU-Pro & $+1.90$ & [$-0.95$, $+5.24$] & 1.00 & $-41.5$ & [$-48.7$, $-33.5$] \\
 & \emph{Macro} & $+1.21$ & [$+0.49$, $+2.00$] & 0.001 & $-39.4$ & [$-41.5$, $-37.3$] \\
\bottomrule
\end{tabular}
\caption{Paired comparison of \romcsc{} with vanilla decoding in all 25 settings and the five per-backbone macro-averages. Point differences follow Table~\ref{tab:main_results_full} (macro length: change in macro-mean tokens); intervals and Holm-corrected $p$-values come from the paired problem-level bootstrap of App.~\ref{sec:stats_app}. Summary: the length reduction is significant in 25/25 settings; accuracy is significantly higher in 14 and significantly lower in none. The $p$-values of the five macro rows are corrected among themselves. With $B{=}20{,}000$, a value below the resolution of the bootstrap is printed as $<$0.003 (25 settings) or $<$0.001 (five macro rows).}
\label{tab:paired}
\end{table*}

\paragraph{\revhead{Uncertainty of the Pareto front.}} \rev{We assess the stability of Pareto-front membership with the same bootstrap. Within each setting, every replicate uses the same resampled problems to recompute all methods' accuracy and mean length. We then record whether \romcsc{} remains non-dominated: no other method is at least as accurate and at least as short, with a strict improvement in at least one metric. Sec.~\ref{sec:baseline_pareto} reports the resulting probabilities of front membership.}

\subsection{Additional Observations}
\label{sec:extra_obs}

\paragraph{Verbosity and gains.}
\romcsc{} delivers its largest benefit on the backbone with the most redundant continuation. Summing vanilla's mean output lengths across the five benchmarks identifies Gemma-4-12B as the most verbose backbone (24.3k tokens); it also achieves the largest macro-average accuracy gain ($+5.47$ points). DS-Llama-8B is the least verbose (6.7k) and gains $+1.21$ points, making Gemma's gain $4.5\times$ larger. On AIME25, the hardest task, \romcsc{} stays within $\pm$0.7 accuracy points of vanilla while reducing output length by 34--43\%.

\paragraph{Method--model interactions.}
Several baselines show strong model dependence. PUMA-RD reduces accuracy on Qwen models ($-13.0$ points on Qwen3-8B/MATH500 and $-10.0$ on Qwen3-14B) but has little adverse effect on the R1 distills. On Gemma-4-12B, it is the only method whose macro-average accuracy falls below vanilla's ($-1.36$). On Qwen3-8B, TERMINATOR dominates REFRAIN on MATH500 and GSM8K, while REFRAIN dominates TERMINATOR on the other three benchmarks. TERMINATOR and NEAT produce outputs 115--118\% as long as vanilla's on AIME25. \rev{This additional length comes from their two-segment runtimes, which generate their own traces. TERMINATOR triggers through a token-level head and NEAT through exit neurons, so neither uses probe decoding. When no trigger occurs, the reasoning segment reaches the shared budget, after which an exit phrase introduces a separately budgeted answer segment. The total output can therefore exceed vanilla's.} Vanilla provides a strong accuracy reference on difficult mathematical tasks. Across MATH500 and AIME25 on the Qwen and DS-R1-32B backbones, Dynasor on DS-R1-32B/MATH500 is the only compression baseline to exceed vanilla accuracy ($+0.67$ points, with shorter output). For \romcsc{}, the three point estimates below vanilla---MATH500 and AIME25 on Qwen3-14B, and AIME25 on DS-R1-32B---differ by $-0.33$ to $-0.67$ points.

\paragraph{Benchmark resolution.}
AIME25 contains 30 problems, with accuracy measured in increments of 1/300; ties among methods reflect this resolution. On GPQA-Diamond, Gemma's accuracy spans 28.8--36.4 against a 25\% four-choice chance level. MMLU-Pro shows the largest gains (up to $+9.52$ on Gemma), with confidence-interval half-widths of $\pm$7.4--9.8 on its 70-question split.

\section[Operating Curves, Early-Stop Rates, pass@k, and Evaluation Scope]{\revhead{Operating Curves, Early-Stop Rates, pass@$k$, and Evaluation Scope}}
\label{sec:more_comparisons}

\subsection{\revhead{Operating-Point Sweeps}}
\label{sec:sweep_app}

\rev{The sweeps in Sec.~\ref{sec:matched_budget} (Fig.~\ref{fig:controls}b) evaluate Qwen3-8B on the main protocol's 100 MATH500 evaluation problems. We vary ROM's decision threshold over $\tau\in\{0.3,0.4,\dots,0.9\}$, Dynasor's (stability threshold, probe chunk size) over $\{(2,64),\allowbreak(3,64),\allowbreak(4,64),\allowbreak(3,32),\allowbreak(3,128)\}$, and EAT's variance threshold over $\delta\in\{0.005,\allowbreak0.01,\allowbreak0.02,\allowbreak0.05,\allowbreak0.1,\allowbreak0.2,\allowbreak0.3,\allowbreak0.5\}$. EAT's four values below 0.1 extend its curve toward longer outputs. Each baseline point is compared with ROM at the same mean output length, obtained by linear interpolation along ROM's threshold curve.}

\subsection{\revhead{Trigger and Early-Stop Rates}}
\label{sec:trigger_app}

\revnextfloat
\begin{table}[t]
\centering
\footnotesize
\setlength{\tabcolsep}{5pt}
\begin{tabular}{@{}lccc@{}}
\toprule
\textbf{Method} & \textbf{Trigger} & \textbf{Early stop} & \textbf{Median cut} \\
\midrule
EAT          & 85.0 & 82.3 & 0.36 \\
RCPD         & 90.0 & 0.0 & 1.00 \\
RPDM         & 3.7 & 3.7 & 0.67 \\
NEAT         & 91.7 & 55.7 & 0.88 \\
TERMINATOR   & 63.0 & 60.3 & 0.27 \\
REFRAIN      & 65.7 & 57.3 & 0.65 \\
Dynasor      & 83.3 & 25.7 & 1.01 \\
PMA          & 77.3 & 74.7 & 0.33 \\
PUMA-RD      & 78.0 & 75.3 & 0.57 \\
\rom{} (w/o CSC) & 82.3 & 75.7 & 0.44 \\
\rowcolor{gray!20} \romcsc{} & 89.7 & 82.7 & 0.41 \\
\bottomrule
\end{tabular}
\caption{Trigger and early-stop rates (\%) on Qwen3-8B/MATH500. \emph{Trigger}: fraction of samples on which the stopping rule fires. \emph{Early stop}: fraction on which it fires before 95\% of the corresponding vanilla reasoning length. \emph{Median cut}: median trigger position relative to that length. All rows use one main-protocol run with 100 evaluation problems and $n{=}3$ samples each, yielding 300 samples per method. NEAT, REFRAIN, and TERMINATOR generate their own 300 traces; their trigger positions are measured against the problem-mean vanilla reasoning length.}
\label{tab:trigger}
\end{table}

\rev{Table~\ref{tab:trigger} separates triggers that occur while reasoning is still underway from those that occur near or after its completion, using the rates defined in Sec.~\ref{sec:trigger_rates}. On Qwen3-8B/MATH500, the 57.7-percentage-point gap between Dynasor's trigger and early-stop rates corresponds to triggers within the final 5\% of vanilla reasoning or later. Nearly all of \romcsc{}'s triggers remove reasoning. Across the 25 settings, \romcsc{} stops early on 52--97\% of samples, with median cut positions at 0.14--0.53 of vanilla reasoning length. Dynasor's early-stop rate ranges from 2\% (Qwen3-8B/AIME25) to 78\% (Gemma-4-12B/MATH500). Its early stops concentrate on traces in which vanilla continues well beyond reaching an answer, and are rare on problems the model cannot solve.}

\subsection{\revhead{avg@16 and pass@16}}

\revnextfloat
\begin{table}[t]
\centering
\footnotesize
\setlength{\tabcolsep}{4pt}
\begin{tabular}{@{}llcc@{}}
\toprule
\textbf{Setting} & \textbf{Method} & \textbf{avg@16} & \textbf{pass@16} \\
\midrule
Qwen3-8B     & Vanilla    & 90.6 & 98.0 \\
MATH500      & Dynasor    & 88.2 & 97.0 \\
             & EAT        & 83.9 & 94.0 \\
             & \romcsc{}  & 90.7 & 98.0 \\
\midrule
Qwen3-8B     & Vanilla    & 26.5 & 46.7 \\
AIME25       & Dynasor    & 26.2 & 43.3 \\
             & EAT        & 25.4 & 43.3 \\
             & \romcsc{}  & 26.2 & 46.7 \\
\midrule
Qwen3-8B     & Vanilla    & 77.1 & 91.4 \\
MMLU-Pro     & Dynasor    & 79.8 & 92.9 \\
             & EAT        & 76.3 & 91.4 \\
             & \romcsc{}  & 79.3 & 92.9 \\
\midrule
Gemma-4-12B  & Vanilla    & 77.9 & 93.0 \\
MATH500      & Dynasor    & 83.7 & 96.0 \\
             & EAT        & 77.8 & 93.0 \\
             & \romcsc{}  & 84.6 & 97.0 \\
\midrule
Gemma-4-12B  & Vanilla    & 32.9 & 50.0 \\
AIME25       & Dynasor    & 32.4 & 50.0 \\
             & EAT        & 31.9 & 50.0 \\
             & \romcsc{}  & 33.1 & 50.0 \\
\midrule
Gemma-4-12B  & Vanilla    & 64.9 & 87.1 \\
MMLU-Pro     & Dynasor    & 72.9 & 91.4 \\
             & EAT        & 70.1 & 90.0 \\
             & \romcsc{}  & 74.2 & 92.9 \\
\bottomrule
\end{tabular}
\caption{Accuracy with $n{=}16$ samples per problem (\%). avg@16 measures mean per-sample accuracy; pass@16 estimates the probability of obtaining at least one correct answer among 16 samples. We use the unbiased pass@16 estimator of \citet{chen2021evaluating} with $n{=}16$ samples per problem. When $n{=}k$, it equals the fraction of problems with at least one correct sample and is therefore a multiple of $1/N$ for $N$ problems (one AIME25 problem corresponds to 3.3 points).}
\label{tab:passk}
\end{table}

\rev{Table~\ref{tab:passk} evaluates vanilla, the two most competitive baselines, and \romcsc{} in six settings with $n{=}16$, keeping the sampling parameters unchanged. For every method, avg@16 is within 1.1 points of the corresponding result in Table~\ref{tab:main_results}. The two metrics capture complementary aspects of performance: avg@$k$ measures per-sample accuracy, while pass@$k$ measures the ability to reach a correct answer across multiple samples. \romcsc{} matches or exceeds vanilla's pass@16 in every setting, with differences ranging from 0 to $+5.7$ points. It ranks highest among the four methods in two settings and ties for highest in the other four. These results show that ROM's gains persist under multi-sample evaluation while preserving the ability to reach correct answers across samples.}

\subsection{\revhead{Evaluation-Set Size}}
\label{sec:scope_app}

\revnextfloat\begin{table}[t]
\centering
\footnotesize
\setlength{\tabcolsep}{3.2pt}
\begin{tabular}{@{}lcccc@{}}
\toprule
 & \textbf{Acc.} & \textbf{Tokens} & \textbf{Cap} & \textbf{Trig.} \\
\midrule
Vanilla   & 67.2 & 2{,}909 & 3.7 & -- \\
\romcsc{} & 67.7 & 1{,}106 & 0.4 & 98.4 \\
$\Delta$ acc. & \multicolumn{4}{l}{$+0.51$ [$-0.20$, $+1.22$]} \\
$\Delta$ tokens & \multicolumn{4}{l}{$-62.0$ [$-63.5$, $-60.5$]} \\
\bottomrule
\end{tabular}
\caption{MMLU-Pro on 980 test questions with Qwen3-8B ($n{=}3$; 2{,}940 samples). Accuracy, \emph{Cap} (fraction of generations reaching the $8{,}192$-token budget), and \emph{Trig.} (fraction of samples triggering \romcsc{}) are percentages. $\Delta$ denotes \romcsc{} minus vanilla: accuracy in points and relative output-length change in \%, with 95\% paired-bootstrap intervals.}
\label{tab:scope}
\end{table}

\rev{The main protocol uses MMLU-Pro's 70-question validation split, comprising five few-shot exemplars from each of 14 categories. We extend the evaluation to 980 test questions with Qwen3-8B, comparing vanilla decoding with \romcsc{}. We randomly select 70 questions per category (seed 46), yielding a set disjoint from the validation split and fourteen times its size. The detector head, threshold $0.5$, backtracing, answer cue, grader, and decoding parameters remain fixed, and the question identifiers are released with the code. On this larger set, \romcsc{} preserves accuracy while removing 62.0\% of the tokens (Table~\ref{tab:scope}).}

\section{\revhead{Cue and Matched-Length Controls}}
\label{sec:cue_app}

\revnextfloat
\begin{table}[t]
\centering
\footnotesize
\setlength{\tabcolsep}{3.4pt}
\begin{tabular}{@{}llccccc@{}}
\toprule
 & & & \textbf{Cue} & & \multicolumn{2}{c}{\textbf{Blind cut}} \\
\cmidrule(l){6-7}
 & & \textbf{Van.} & \textbf{only} & \textbf{ROM} & \textbf{Fixed} & \textbf{Rand.} \\
\midrule
Qwen3    & MATH500  & 90.3 & 91.3 & 90.7 & 84.0 & 81.0 \\
-8B      & AIME25   & 27.0 & 27.0 & 27.3 & 18.3 & 15.3 \\
         & MMLU-Pro & 76.7 & 76.2 & 79.0 & 71.9 & 69.5 \\
\midrule
Gemma    & MATH500  & 78.3 & 79.7 & 85.0 & 69.0 & 63.7 \\
-4-12B   & AIME25   & 32.3 & 32.0 & 33.0 & 17.3 & 12.7 \\
         & MMLU-Pro & 64.3 & 64.8 & 73.8 & 58.6 & 55.7 \\
\bottomrule
\end{tabular}
\caption{Separating the effects of the answer cue and cut position (accuracy, \%; ROM denotes \romcsc{}). \emph{Cue only}: full vanilla reasoning followed by ROM's final-answer cue. \emph{Fixed} / \emph{Rand.}: content-blind cuts matching ROM's reasoning retention, with identical backtracing, cue, and grading procedures (Rand.: three seeds).}
\label{tab:cue_control}
\end{table}

\paragraph{\revhead{Conditions.}} \rev{Let $\ell_j$ denote the reasoning length of vanilla trace $j$. For each setting, we compute \romcsc{}'s reasoning-retention ratio $\rho$ as the total retained tokens divided by the total reasoning tokens, counting untriggered traces as retained in full. The conditions in Table~\ref{tab:cue_control} share ROM's backtracing rule, final-answer cue, continuation sampling, and grader, and differ in where the reasoning is cut.}
\begin{itemize}[leftmargin=*,itemsep=1pt,topsep=2pt]
\item \rev{\emph{Cue only.} The full vanilla reasoning trace is retained. We append the cue where the model closes its reasoning or, for traces that exhaust the budget, at the end of the budget-limited trace.}
\item \rev{\emph{Fixed.} Every trace is cut at $\lfloor \rho\,\ell_j\rfloor$.}
\item \rev{\emph{Random.} Trace $j$ is cut at $\lfloor u_j \ell_j\rfloor$ with $u_j\sim\mathcal{U}[\max(0,2\rho{-}1),\,\min(1,2\rho)]$, an interval centered on $\rho$; results are averaged over three seeds (46, 47, 48).}
\end{itemize}
\rev{Both content-blind controls use ROM's retention ratio to set cut positions without inspecting the reasoning content.}

\paragraph{\revhead{Results.}} \rev{Two observations complement the results in Sec.~\ref{sec:cue_control}. First, the cue-only control also elicits answers from traces that exhaust the budget. Its small effect shows that completing unfinished generations alone does not explain ROM's gains. Second, \romcsc{} exceeds the content-blind controls by 6.7--16.0 accuracy points for fixed cuts and 9.5--21.3 points for random cuts. Relative to vanilla, the content-blind controls incur their largest accuracy losses on AIME25 and their smallest on MMLU-Pro for both backbones. This pattern highlights the value of selecting a suitable boundary in long mathematical derivations.}

\section{Ablation Details}
\label{sec:ablation_details}

\subsection{\revhead{Isolating the Contribution of CSC}}
\label{sec:csc_app}

\rev{This appendix details the three Qwen3-8B analyses in Sec.~\ref{sec:csc_causal} that isolate the contribution of CSC's trajectory content.}

\paragraph{\revhead{Size controls.}} \rev{We retrain the head under the same protocol on the two training sets in Table~\ref{tab:csc_variants}. The first retains both CSC components and is downsampled to 568 distinct samples, matching the unaugmented set. The second disables the counterfactual branch in Alg.~\ref{alg:augmentation}, so each response is processed as if no incorrect attempt were available. It contains 1{,}336 distinct samples drawn from original and resampled correct-solution views, upsampled to match the full CSC set's size. The resulting output length is comparable to that of \rom{} without CSC, isolating the benefit of the counterfactual component.}

\paragraph{\revhead{Attempt-index shortcut.}} \rev{We measure reliance on attempt index by testing whether detector scores distinguish later attempts after accounting for token position. We regress token position out of the detector score and measure how well the residual distinguishes tokens in attempts with index $\geq 2$ from tokens in first attempts. The analysis uses tokens through the end of the first correct solution from the 145 held-out Qwen3-8B traces described below: 15 with $k^*{\geq}2$ and 130 controls. All selected tokens have the efficient label, allowing us to vary attempt index while holding the supervision target fixed.}

\paragraph{\revhead{Natural wrong$\rightarrow$correct traces.}} \rev{We apply the segmentation and correctness pipeline in App.~\ref{sec:label_prompts} to held-out Qwen3-8B vanilla traces. Selecting traces that begin with an incorrect attempt and later reach a correct solution ($k^*{\geq}2$) yields 15 traces, which we compare with 130 control traces without self-correction. A cut is \emph{premature} if it occurs before the end of the first correct solution.}

\subsection{Detector and Control}

All configurations in Table~\ref{tab:ablations} use the main evaluation protocol with $300$ traces each. Truncation, backtracing, and continuation procedures are held fixed except for the component being ablated, enabling direct comparison across configurations and with vanilla decoding in Table~\ref{tab:main_results}.

\paragraph{Detector architecture.}
The linear head replaces the recurrent cell while retaining the same attention features, training data, and training protocol. The confidence moving-average trigger (window $16$) fires on locally confident derivation steps at both tested thresholds, reducing accuracy by $20$--$23$ points.

\paragraph{Backtracing.}
Backtracing preserves a coherent prefix when a trigger falls within a sentence or derivation. Removing it increases output length by $+6.7\%$ (Table~\ref{tab:ablations}), consistent with the model reconstructing interrupted steps during regeneration. Accuracy also decreases by $1.0$ point, or three of $300$ traces. Together, these results support backtracing as a way to maintain reasoning coherence and avoid compensatory generation.

\paragraph{Layer and threshold.}
We probe layers $22$, $32$, and $34$ of the $36$-layer backbone, retraining the detector at each layer. L32, used in the main results, is selected by the fixed depth rule in App.~\ref{sec:head_details}, independently of this sweep. Performance is robust across the probed layers: each removes $47$--$54\%$ of tokens, with accuracy spanning $1.4$ points around vanilla. Length savings also vary smoothly with the trigger threshold.

\section{\revhead{Efficiency Measurement Details}}
\label{sec:efficiency_app}

\paragraph{\revhead{Streaming implementation.}} \rev{The online decoder runs a batch-size-1 loop in Hugging Face Transformers with a dynamic KV cache and \texttt{bfloat16} precision. At each generated token, the detector projects the layer-$L$ hidden state to a query, key, and value, then appends the key and value to its cache. It attends over the cached prefix, including the prompt, updates the recurrent cell, and applies the output layer. The initial state $\mathbf{m}_0$ is computed once from the prompt (App.~\ref{sec:head_details}). When $p_t>0.5$, the decoder backtraces to a valid boundary, crops the backbone and detector caches to the retained prefix, prefills the cue, and resumes answer generation within the same $8{,}192$-token response budget. For Gemma-4, whose sliding-window attention layers cannot be cropped once the window is exceeded, the decoder restores the retained prefix and cue by re-prefilling.}

\paragraph{\revhead{Protocol.}} \rev{We time each backbone on the same 130 problems, taking the first 26 from each benchmark and generating one sample per problem. Vanilla and ROM run back to back on each problem with the same seed and GPU, producing identical tokens up to the trigger. CUDA-event timing covers the full generation call from prompt prefill to the final token, including the detector, backtracing, cache cropping or re-prefill, cue prefill, and regenerated answer. Tokenization and model loading are excluded. We use NVIDIA RTX PRO 6000 Blackwell GPUs (96\,GB) shared with other jobs, pairing every comparison on the same device. This timing protocol uses its own problem subset, sample count, and streaming decoder, so its token totals differ from Table~\ref{tab:main_results}. Within this subset, measured latency reductions are within 2 percentage points of those predicted from decoded-token counts.}

\paragraph{\revhead{Latency.}} \rev{Fig.~\ref{fig:controls}c reports latency reductions for each benchmark. Pooled across the 130 problems, mean latency falls from 172.0\,s to 107.8\,s on Qwen3-8B ($-37.3\%$) and from 356.8\,s to 175.6\,s on Gemma-4-12B ($-50.8\%$). For Qwen3-8B, the pooled reduction is below the macro-average ($46.1\%$) because AIME25 contributes the longest runtimes and has the smallest relative reduction, from 277.8\,s to 221.8\,s ($-20.1\%$). On Gemma-4-12B, AIME25 latency falls from 516.0\,s to 240.9\,s ($-53.3\%$).}

\revnextfloat
\begin{table}[t]
\centering
\footnotesize
\setlength{\tabcolsep}{3.2pt}
\begin{tabular}{@{}lccccc@{}}
\toprule
\textbf{Prefix (tokens)} & \textbf{1K} & \textbf{2K} & \textbf{4K} & \textbf{8K} & \textbf{16K} \\
\midrule
\multicolumn{6}{@{}l}{\emph{Qwen3-8B}} \\
\quad Head, ms/step     & 0.108 & 0.106 & 0.113 & 0.128 & 0.162 \\
\quad Share of step (\%) & 0.30 & 0.31 & 0.30 & 0.32 & 0.42 \\
\midrule
\multicolumn{6}{@{}l}{\emph{Gemma-4-12B}} \\
\quad Head, ms/step     & 0.104 & 0.104 & 0.110 & 0.123 & 0.147 \\
\quad Share of step (\%) & 0.16 & 0.18 & 0.18 & 0.21 & 0.24 \\
\bottomrule
\end{tabular}
\caption{Detector cost per decoding step across prefix lengths. \emph{Share of step}: detector overhead relative to the rest of the decoding step, measured in paired runs. Prefixes are taken from real AIME25 traces.}
\label{tab:prefix_sweep}
\end{table}

\paragraph{\revhead{Detector cost vs.\ prefix length.}} \rev{Table~\ref{tab:prefix_sweep} reports detector cost at five prefix lengths. Both the detector and the backbone read caches that grow by a fixed amount per token: one key and one value with $1{,}024$ dimensions in \texttt{bfloat16} for the detector, and 36 layers $\times$ 8 key--value heads $\times$ 128 dimensions $\times$ 2 tensors $\times$ 2 bytes for Qwen3-8B. When cache reads dominate, the ratio of these sizes (Sec.~\ref{sec:latency}) determines the detector's relative cache-read cost. Weight reads also favor the detector: it reads approximately 38\,MiB per step, compared with 15.3\,GiB for the backbone (0.24\%). The measured runtime share starts near this weight-read ratio and increases slowly with prefix length, consistent with the growing contribution of cache reads. Combining weight and cache bytes read per step gives a relative memory-read cost of approximately 0.6\% on Qwen3-8B and 0.4\% on Gemma-4-12B at a 16K-token prefix.}

\paragraph{\revhead{Memory.}} \rev{The microbenchmark in Table~\ref{tab:prefix_sweep} isolates the detector's memory footprint. Across the five prefix lengths, it adds 51--146\,MiB of peak memory on Qwen3-8B and 66--150\,MiB on Gemma-4-12B. The end-to-end measurements in Sec.~\ref{sec:latency} report the difference in median peak memory across complete generations.}

\section{Sample-Level Analysis and Case Studies}
\label{sec:case_eat_rom}

\paragraph{Agreement with vanilla.}
On MMLU-Pro with Qwen3-8B (70 problems with three samples each), vanilla decoding and ROM$_{\text{CSC}}$ are both correct on 156 of the 210 samples (74.3\%) and both incorrect on 39 (18.6\%). ROM$_{\text{CSC}}$ corrects 10 of vanilla's 49 incorrect samples (20.4\%), while 5 of the 161 originally correct samples become incorrect (3.1\%). This yields a net gain of 5 correct samples ($+2.38$ accuracy points, from 76.67\% to 79.05\%), matching the Qwen3-8B/MMLU-Pro results in Table~\ref{tab:main_results}. \rev{The cue-only control changes accuracy by $-0.5$ points in this setting (Table~\ref{tab:cue_control}), supporting the contribution of boundary selection to this gain.}

\paragraph{Entropy-based vs.\ boundary-based stopping.}
Table~\ref{tab:case_eat} summarizes two cases that illustrate how learned boundary detection handles distinct failure modes of entropy-based stopping (Sec.~\ref{sec:entropy_comparison}). Both use Qwen3-8B and are selected from aligned outputs in which EAT answers incorrectly and ROM$_{\text{CSC}}$ answers correctly.

\begin{table}[t]
\centering
\footnotesize
\setlength{\tabcolsep}{3pt}
\begin{tabular}{@{}lcrrl@{}}
\toprule
\textbf{Method} & \textbf{Ans.} & \textbf{Reason.} & \textbf{Resp.} & \textbf{Stop signal} \\
\midrule
\multicolumn{5}{@{}l}{\emph{Case 1: confidently wrong} (gold D)} \\
\cmidrule{1-5}
\quad EAT      & I~\xmark          & 1{,}570          & 2{,}194          & exit, $H{=}7.5{\times}10^{-5}$ \\
\quad \romcsc{} & \textbf{D}~\cmark & \textbf{906}     & \textbf{1{,}641} & cut at 886 \\
\midrule
\multicolumn{5}{@{}l}{\emph{Case 2: correct but uncertain} (gold F)} \\
\cmidrule{1-5}
\quad EAT      & J~\xmark          & 6{,}802          & 7{,}734          & no exit, $\bar{H}{=}1.61$ \\
\quad \romcsc{} & \textbf{F}~\cmark & \textbf{2{,}898} & \textbf{3{,}938} & cut at 2{,}878 \\
\bottomrule
\end{tabular}
\caption{Two MMLU-Pro cases (Qwen3-8B) illustrating the benefits of boundary-based stopping. EAT exits early when entropy collapses onto an incorrect option (Case 1) and never exits when entropy remains unsettled (Case 2); \romcsc{} preserves the correct answer in both. Lengths count reasoning\,/\,response tokens.}
\label{tab:case_eat}
\end{table}

\subsection{Case 1: Confidently Wrong}

\textbf{Question.} A mutation in a bacterial enzyme changed a previously polar amino acid into a nonpolar amino acid. This amino acid was located at a site distant from the enzyme's active site. How might this mutation alter the enzyme's substrate specificity? The correct option is D: by changing the shape of the protein. A strong distractor is I: a change away from the active site cannot alter specificity.

\noindent \textbf{Analysis.}
The model initially identifies the correct mechanism: a distant polar-to-nonpolar mutation can perturb protein folding and indirectly change the active-site conformation. Continued reasoning replaces this explanation with the false absolute rule that only active-site mutations can affect specificity. EAT exits at checkpoint 21 of 44 with near-zero post-\texttt{</think>} entropy, reflecting high confidence in an incorrect answer. ROM$_{\text{CSC}}$ intervenes before the answer drifts and regenerates the correct option, D.

\transcriptcaption{Abridged EAT trace for the confidently wrong case.}
\begin{transcriptbox}{EAT Trace (abridged)}
\begin{scriptsize}
\begin{verbatim}
[...early reasoning...]
even though the mutation is away from the active site, it
could indirectly affect the active site's conformation,
thereby changing the substrate specificity.

[...continued deliberation...]
If the amino acid is not in the active site, then the
mutation doesn't directly affect the active site...
I think I is the answer.

[EAT exit: entropy = 7.5e-5]
Final answer: I.
\end{verbatim}
\end{scriptsize}
\end{transcriptbox}

\transcriptcaption{Abridged ROM$_{\text{CSC}}$ trace for the confidently wrong case.}
\begin{transcriptbox}{ROM$_{\text{CSC}}$ Trace (abridged)}
\begin{scriptsize}
\begin{verbatim}
[...same early reasoning...]
If the protein's shape changes, even if the active site is
not directly affected, the overall structure might alter
the active site's conformation.

So, the answer might be D: By changing the shape of the
protein.

[ROM cuts at token 886]
Final answer: D.
\end{verbatim}
\end{scriptsize}
\end{transcriptbox}

\subsection{Case 2: Correct but Uncertain}

\textbf{Question.} A state prohibits disposal of nuclear waste within the state. A local disposal company had contracts with out-of-state firms and can no longer perform them. Assuming standing, what is the strongest constitutional ground for challenging the state law? The correct option is F: the Commerce Clause. The distractor selected by EAT is J: N/A.

\noindent \textbf{Analysis.}
Here, ROM identifies sufficient reasoning even while the model's confidence remains unsettled. The model repeatedly identifies the Commerce Clause as the relevant constitutional basis, then reconsiders previously eliminated alternatives. EAT never exits because the entropy trajectory remains unstable. Further elimination leads the model to reject the Commerce Clause and choose N/A. ROM$_{\text{CSC}}$ detects the boundary after the useful legal analysis is complete and preserves the correct conclusion before this reversal.

\transcriptcaption{Abridged EAT trace for the correct-but-uncertain case.}
\begin{transcriptbox}{EAT Trace (abridged)}
\begin{scriptsize}
\begin{verbatim}
Option F: Commerce Clause... the man's contracts are with
out-of-state firms. So the disposal of nuclear waste in the
state would involve interstate commerce.

[...many rounds of uncertainty...]
So perhaps the answer is F. Let me think again.
... But I'm not sure.

[...continued elimination after no EAT exit...]
The law does not violate the Commerce Clause...
J. N/A is the only remaining option.
\end{verbatim}
\end{scriptsize}
\end{transcriptbox}

\transcriptcaption{Abridged ROM$_{\text{CSC}}$ trace for the correct-but-uncertain case.}
\begin{transcriptbox}{ROM$_{\text{CSC}}$ Trace (abridged)}
\begin{scriptsize}
\begin{verbatim}
Option F: Commerce Clause... the state law could be
challenged under the Commerce Clause because it affects
interstate commerce.

[...before the later over-elimination...]
So F is the answer.

[ROM cuts at token 2,878]
Final answer: F.
\end{verbatim}
\end{scriptsize}
\end{transcriptbox}

\end{document}